\documentclass{article}
\usepackage{naturetex}
\usepackage{tabularx}
\usepackage{amsmath}
\usepackage{multirow}
\usepackage{nicematrix}
\usepackage{csquotes}

\newcolumntype{C}{>{\centering\arraybackslash}X}
\usepackage{caption}
\usepackage{rotating}
\usepackage{float}
\usepackage{booktabs}
\usepackage{longtable}
\usepackage{makecell}
\usepackage{changepage}

\usepackage{chngcntr}

\usepackage{hyperref}

\begin{document}

\maketitle

\newpage
{\setstretch{1.0}
	\section*{Abstract}
Large language models (LLMs) have shown promise in supporting medical diagnosis, with prompting-based methods offering a flexible and deployable means of capability enhancement. 
However, existing prompt engineering and multi-agent approaches often focus on optimizing single inferences, paying less attention to the accumulation of reusable experience from clinical practice, constraining their real-world applicability.
To address this, this study proposes a novel Multi-Agent Clinical Diagnosis (MACD) framework, which allows LLMs to self-learn clinical knowledge via a multi-agent pipeline that summarizes, refines, and applies diagnostic insights, mirroring the professional development of human physicians.
We further extend it to a MACD-human collaborative workflow, where multiple LLM-based diagnostician agents engage in iterative consultations, supported by a judge agent and human oversight for cases where agreement is not reached.
The MIMIC-MACD cohort comprising 4,390 real-world patient cases across seven diseases is constructed, including 1,314 cases for knowledge learning and 3,076 held-out cases for evaluation.
Across diverse open-weight LLMs (Llama-3.1 8B/70B, DeepSeek-R1-Distill-Llama 70B),  { MACD significantly improves primary diagnostic accuracy, achieving an average improvement of 11.6 percentage points (and up to 22.4 percentage points) over established authoritative knowledge, while narrowing the performance gap between open-weight models and state-of-the-art LLMs.} 
Furthermore, the MACD-human workflow yields an 18.3-percentage-point improvement over physician-only diagnosis  { on text-only vignettes}, demonstrating the synergistic potential of human-AI collaboration.
Notably, the self-learned knowledge exhibits strong cross-model stability, transferability across LLMs, and capacity for model-specific personalization. The system also generates traceable diagnostic rationales, enhancing transparency and explainability.
This work thus presents a scalable self-learning paradigm that bridges the gap between the intrinsic knowledge of LLMs and the demands of real-world clinical practice, advancing towards a reliable, interpretable, and deployable AI-assisted diagnosis.
}

\newpage
\section*{Introduction}

As societal development accelerates, the unequal distribution of medical resources remains a critical global challenge constrained by the prohibitive costs of traditional infrastructure~\cite{custers2015thirty,zelikman2024star}. Against this backdrop, the rapid advancement of Large Language Model (LLM) technology offers transformative potential to address the shortage of medical expertise and the uneven distribution of healthcare resources~\cite{chang2024survey, gomez2024artificial,delourme2024measured}. In real-world medical settings, physicians typically confront open-ended diagnoses, whereas current LLM-based solutions predominantly focus on question-answering tasks~\cite{jin2019pubmedqa,jin2021disease,thirunavukarasu2023large}, which rely on simplified scenarios that do not capture the complexity of open-ended diagnoses~\cite{singhal2025toward,chen2023meditron,hager2024evaluation}. Consequently, a significant performance gap emerges when LLMs encounter clinical data from real-world patients~\cite{gaber2025evaluating,bedi2024systematic}.

While scaling model capabilities and specialized post-training effectively enhance the medical performance of LLMs, the substantial computational resources required for these approaches impede their deployment in resource-constrained grassroots settings~\cite{qiu2025quantifying,kim2025small,chen2024huatuogpt}. Conversely, cost-effective prompt engineering strategies, such as chain-of-thought and few-shot, enhance diagnostic capabilities by guiding the reasoning of LLMs~\cite{savage2024diagnostic,guo2025structured,zhao2024effective,sahoo2024systematic}. However, these methods perform each diagnostic inference independently, failing to leverage reusable clinical experience.
In real-world medical settings, physicians efficiently diagnose by complementing standard guidelines with their own accumulated experiential knowledge~\cite{custers2015thirty, thomas1996guidelines}. Inspired by this, integrating the ability to build upon and reuse past diagnostic knowledge is, therefore, a critical next step for clinical AI systems to mature.

\begin{figure}[htbp]
	\begin{center}
		\includegraphics[width=1\textwidth]{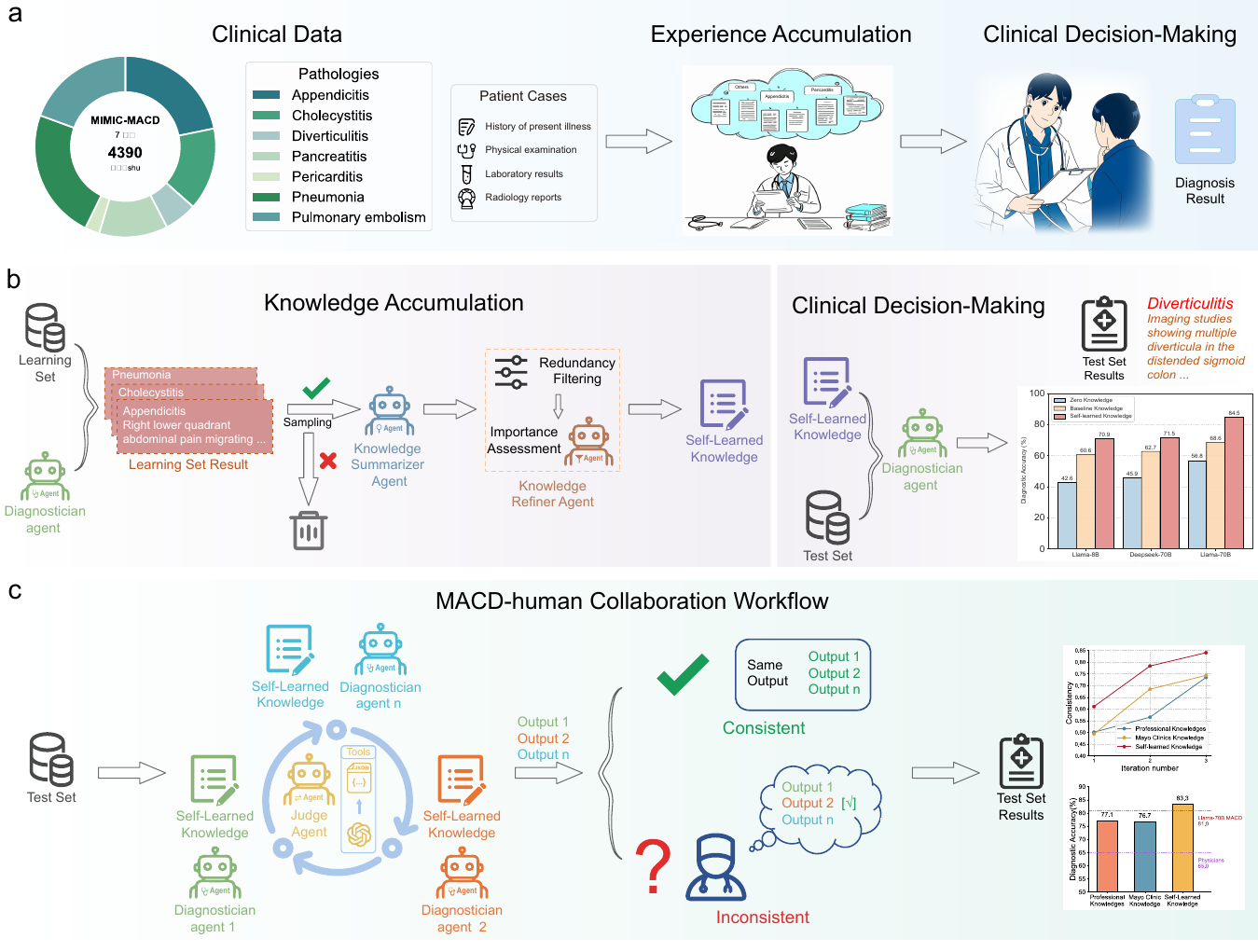}
	\end{center}
	\caption{\textbf{Overview of the MACD framework.} (a) The core idea of MACD is to emulate the professional development of a physician by enabling the LLM to autonomously acquire, distill, and internalize clinical knowledge from real-world diagnostic cases over time. (b) MACD is formed with a knowledge summarizer agent that identifies and extracts salient diagnostic insights from historical cases, a knowledge refiner agent that consolidates and integrates these insights into a structured, evolving knowledge memory, and a diagnostician agent that leverages this curated experience to inform and improve diagnostic reasoning. (c) A workflow comprising multiple MACD diagnostician agents based on diverse LLMs, called the MACD-human collaboration workflow. The three diagnostic agents jointly discuss the cases. The final output is provided either when agreement is reached or, if the maximum number of discussions is reached without consensus, the final diagnosis is made by human physicians. The figure is drawn by the authors using Microsoft PowerPoint, with icons from its built-in icon library.}
	\label{fig:fig1}
\end{figure}

Here, we propose a novel LLM-based multi-agent clinical diagnosis (MACD) framework that enhances diagnostic accuracy via the systematic accumulation of reusable clinical experience. As illustrated in Fig.~\ref{fig:fig1}a, the core idea is to emulate the professional development of a physician by enabling the LLM to autonomously acquire, distill, and internalize clinical knowledge from real-world diagnostic cases over time. Within the framework, as shown in Fig.~\ref{fig:fig1}b, there is an ensemble of specialized agents performing distinct cognitive roles: a \textit{knowledge summarizer agent} identifies and extracts salient diagnostic insights from historical cases; a \textit{knowledge refiner agent} consolidates and integrates these insights into structured, evolving self-learned knowledge; and a \textit{diagnostician agent} leverages this curated Self-Learned Knowledge as a critical part of the LLM prompt to inform and improve diagnostic reasoning based on the LLM. This architecture not only mirrors the experiential learning trajectory of a medical expert but also establishes a scalable, self-improving paradigm for LLM-assisted diagnosis.

Based on the above MACD framework, this study further develops a MACD-human collaboration workflow. As depicted in Fig.~\ref{fig:fig1}c, the workflow integrates diagnostician agents based on diverse LLMs. These diagnostician agents, all leveraging the core self-learning mechanism, engage in multiple rounds of consultation to exchange opinions and reach consensus before producing a final diagnosis. The system also includes a \textit{judge agent} responsible for verifying the consistency of the diagnosis. For a small fraction of cases where the diagnostician agents fail to reach an agreement, the workflow introduces human physicians to enable human-LLM collaboration: the outputs of the diagnostician agents serve as decision support, while the final judgment is made by the human physician. In performance evaluations, the collaborative framework significantly outperforms any single agent, validating its scalability and clinical potential.

This study systematically evaluates the effectiveness and generality of the framework, as well as the impact of the base model on its performance.
The key findings of the study are as follows: 
(1) The Self-Learned Knowledge (SLK) generated by the MACD framework is more suitable for LLMs than the Professional Knowledge~\cite {konstantinides20202019, metlay2019diagnosis,torres2017international,asteggiano20152015,leppaniemi20192019,sartelli20202020,pisano20202020,di2016wses,di2020diagnosis} and Mayo Clinic Knowledge~\cite{noauthor_appendicitis_nodate, noauthor_cholecystitis_nodate, noauthor_diverticulitis_nodate, noauthor_pancreatitis_nodate,noauthor_pericarditis_nodate,noauthor_pneumonia_nodate,noauthor_pulmonary_nodate} in clinical diagnosis,  { boosting diagnostic accuracy by an average of 11.6 percentage points and up to 22.4 percentage points.}
 { When equipped with the SLK, the diagnostician agents significantly narrow the diagnostic performance gap between open-weight models and state-of-the-art (SOTA) LLMs (e.g., GPT-5) under identical knowledge and prompting conditions, highlighting the effectiveness of self-learned knowledge in improving diagnostic performance.} Furthermore, the framework establishes a distinct advantage over both prompt engineering strategies (e.g., Chain-of-Thought, Few-Shot) and fine-tuning methods, providing empirical evidence that simulating experience accumulation yields greater diagnostic gains than traditional algorithmic adjustments.
(2) Compared to authoritative knowledge, the SLK demonstrates superior stability, yielding predictable performance improvements that align with models’ intrinsic capabilities (stronger agents obtain higher accuracy, weaker ones lower), and greater transferability, achieving substantial and consistent accuracy gains across diverse LLMs.
(3) The SLK generated by the MACD framework exhibits model-specific preferences and individualized characteristics, primarily manifested in the fact that each model achieves the highest diagnostic accuracy when using knowledge it has generated itself compared to knowledge derived from other models. Simultaneously, the SLK achieves favorable results in disease relevance assessments by human experts, demonstrating promising clinical value.
(4) The SLK is further evaluated to be effective in the MACD-human collaboration workflow. The workflow, through its consultation process, provides human physicians with expanded diagnostic perspectives. The SLK achieves 
the highest diagnostic consensus rate (84.0\%) and 
the highest effective opinion rate (86.3\%) compared to the other two authoritative knowledge sources, indicating its greater capacity to provide physicians with effective diagnostic perspectives. In the subset evaluation where human physicians participate in the consultation process, the SLK still achieves the highest diagnostic accuracy of  { 83.3\%}, outperforming the physicians-only diagnosis accuracy  { on text-only vignettes} by 18.3 percentage points.
(5) The framework enhances explainability by generating a transparent text-based rationale that links the diagnosis to the case and knowledge, ensuring traceable and interpretable reasoning.
In summary, the MACD framework bridges the gap between LLM capabilities and clinical reality by significantly improving diagnostic performance. Its cost-effective, self-learning characteristics make it suitable for grassroots deployment, offering a practical solution to address global healthcare inequality.

\newpage
\section*{Results} \label{results}

\subsection*{The MACD framework is constructed following a human cognition process}

To simulate a physician’s professional growth through the accumulation of real-world diagnostic experience (Fig.~\ref{fig:fig1}a), we propose the MACD framework, which models key cognitive processes such as iterative case summarization, mental model refinement, and knowledge application, as shown in Fig.~\ref{fig:fig1}b. By enabling multiple agents to collaboratively learn from and reason over clinical cases, MACD supports scalable and efficient triage in high-throughput clinical environments. Moreover, the framework facilitates the transfer of diagnostic expertise from tertiary medical centers to primary care settings, helping to alleviate disparities in healthcare resources and clinical decision-making capacity.

The framework is operationalized by a team of three collaborating agents: a knowledge summarizer agent that identifies and extracts salient diagnostic insights from historical cases, a knowledge refiner agent that consolidates and integrates these insights into a structured, evolving knowledge memory, and a diagnostician agent that leverages this experience to inform and improve diagnostic reasoning. For a detailed description, please refer to the `\nameref{Multi-Agents}' section in \nameref{meth}. 
Technically, to satisfy the strict privacy requirements of clinical settings, we adopt three on-premises deployable base LLMs with diverse parameter scales and training strategies: Llama 3.1-8B-Instruct, Llama 3.1-70B-Instruct~\cite{meta2024introducing}, and DeepSeek-R1-Distill-Llama-3.3-70B (DeepSeek-70B)~\cite{guo2025DeepSeek}.
To evaluate the practical efficacy of the Self-Learned Knowledge autonomously generated by the collaboration of the MACD framework, we establish a Baseline Knowledge benchmark. This benchmark represents the average diagnostic accuracy achieved using two authoritative external sources: Professional Knowledge extracted via Gemini 2.5 Pro sourced from institutional standards~\cite{konstantinides20202019, metlay2019diagnosis, torres2017international, asteggiano20152015, leppaniemi20192019, sartelli20202020, pisano20202020, di2016wses, di2020diagnosis} and the Mayo Clinic Knowledge sourced from the official website~\cite{noauthor_appendicitis_nodate, noauthor_cholecystitis_nodate, noauthor_diverticulitis_nodate, noauthor_pancreatitis_nodate,noauthor_pericarditis_nodate,noauthor_pneumonia_nodate,noauthor_pulmonary_nodate}.
We construct the MIMIC-MACD dataset (Fig.~\ref{fig:fig1}a, Supplementary Table 1), which consists of 4,390 cases sourced from the MIMIC-CDM dataset~\cite{hager2024evaluation} and the MIMIC-IV v2.2~\cite{goldberger2000physiobank}, encompassing seven diseases. Each case includes four categories of information: history of present illness, physical examination, laboratory results, and image reports. The dataset is partitioned into a learning set for knowledge self-learning and a test set for evaluation. All self-learning activities are confined to the cases within the learning set.  { All reported experimental results are from the test set, and their performance is characterized using primary diagnostic accuracy with percentage points.}

\subsection*{Self-learned knowledge exhibits superiority in optimizing clinical diagnostics}

The proposed MACD framework demonstrates remarkable effectiveness and generalizability as a `plug-and-play' solution for clinical diagnosis optimization. By validating the framework through extensive comparative experiments, we confirm that the Self-Learned Knowledge generated by the MACD agent team significantly enhances diagnostic accuracy across diverse models. This improvement offers a scalable pathway to upgrade diagnostic capabilities in resource-constrained healthcare settings.

\begin{figure}[htbp]
	\begin{center}
		\includegraphics[width=1\textwidth]{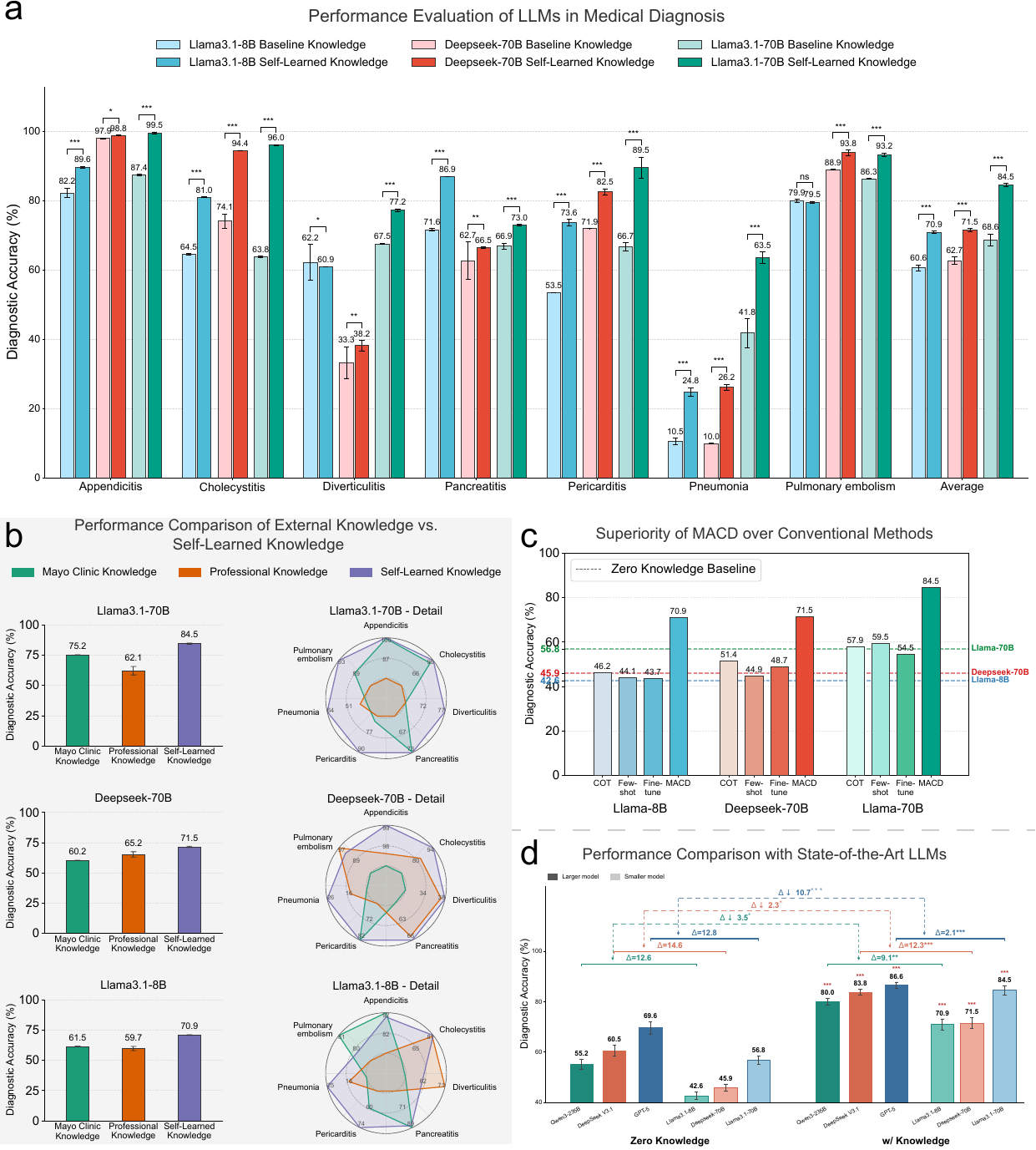}
	\end{center}
	\caption{\textbf{Comparison of diagnostic accuracy between self-learned knowledge and baseline knowledge.} (a) Comparison of diagnostic accuracy across seven diseases, illustrating the consistent improvement of self-learned knowledge over baseline performance. (ns: not significant, *: $p < 0.05$, **: $p < 0.01$, ***: $p < 0.001$) (b) Performance assessment contrasting self-learned knowledge with Mayo Clinic's knowledge and professional knowledge, demonstrating superior diagnostic accuracy. (c) Benchmarking results highlighting the advantage of MACD over established inference methods and resource-intensive fine-tuning.  { (d) Comparative analysis revealing that SLK narrows the performance gap between open-weight models and state-of-the-art LLMs.}}
	\label{fig:fig2}
\end{figure}

We first investigate whether MACD can unlock the latent diagnostic potential of base LLMs through self-learned clinical knowledge and whether this knowledge can outperform authoritative knowledge sources (Fig.~\ref{fig:fig2}a).
Compared with the Authoritative Baseline Knowledge, the Self-Learned Knowledge consistently improves diagnostic accuracy across all evaluated models, yielding gains of +10.3 percentage points for Llama-8B, +8.7 percentage points for DeepSeek-70B, and +15.9 percentage points for Llama-70B, with an average improvement of 11.6 percentage points across seven diseases.

To further assess the quality of the learned knowledge itself, we compare it against human-curated authoritative sources, including Professional Knowledge derived from institutional standards and Mayo Clinic Knowledge (Fig.~\ref{fig:fig2}b). Across models and disease categories, the Self-Learned Knowledge exhibits a substantially stronger guiding effect. In particular, the Llama-70B-based agent outperforms Professional Knowledge by 22.4 percentage points and Mayo Clinic Knowledge by 9.3 percentage points. Even for smaller models such as Llama-8B, MACD achieves improvements of +11.2 percentage points over Professional Knowledge and +9.4 percentage points over Mayo Clinic Knowledge.

We attribute this superiority to the fundamental difference in knowledge generation. While authoritative knowledge is curated specifically for human cognition, Self-Learned Knowledge is distilled directly from real-world clinical cases. Consequently, its content is more attuned to the complexities of actual clinical scenarios and maintains an intrinsic alignment with the model's own inferential patterns, thereby bridging the gap between theoretical knowledge and practical application, suggesting its potential to serve as a reliable automated opinion for rapid triage.

Finally, a detailed analysis (Fig.~\ref{fig:fig2}b, radar graphs) reveals a scaling effect in agent collaboration: stronger base models, such as Llama-70B, derive increasingly greater benefits from self-learned knowledge, consistently widening their performance margin over human-authored guidelines across all diseases.

We next benchmark MACD against established methodological paradigms to determine whether its gains extend beyond conventional prompting and model adaptation strategies.
Compared to the intrinsic capabilities of the base models (Zero Knowledge), the MACD framework consistently delivers substantial performance gains, exceeding 25 percentage points across all tested configurations (Fig.~\ref{fig:fig2}c, Supplementary Figure 1). Crucially, from a resource allocation perspective, MACD proves superior to both conventional prompt engineering strategies and computationally expensive training methods (Fig.~\ref{fig:fig2}c, Supplementary Table 3). We conducted a comparative analysis against Chain-of-Thought (CoT), Few-shot learning, and LoRA fine-tuning trained on the same learning set. The results are striking: the MACD framework (e.g., achieving 84.5\% on Llama-70B) significantly outperforms not only inference-time techniques like CoT (e.g., 57.9\%) but also the resource-intensive Fine-tuning (e.g., 54.5\%) across all three models. Ultimately, MACD demonstrates superior diagnostic precision over conventional methods without the need for cumbersome post-training, offering a scalable, cost-effective solution for optimizing clinical pathways.

{ We further examine whether Self-Learned Knowledge (SLK) can narrow the performance gap between open-weight models and state-of-the-art (SOTA) LLMs under controlled prompting and knowledge conditions (Fig.~\ref{fig:fig2}d).
Under the identical zero-knowledge condition, SOTA models achieve accuracies of 55.2\% (Qwen3-235B), 60.5\% (DeepSeek V3.1), and 69.6\% (GPT-5), whereas open-weight models achieve 42.6\% (Llama3.1-8B), 45.9\% (DeepSeek-70B), and 56.8\% (Llama3.1-70B), respectively, revealing a similar low-to-high capability ordering across the two groups. We therefore compare models at corresponding relative capability levels, yielding zero-knowledge gaps of 12.6, 14.6, and 12.8 percentage points for Qwen3-235B versus Llama3.1-8B, DeepSeek V3.1 versus DeepSeek-70B, and GPT-5 versus Llama3.1-70B, respectively. When both models within each comparison are subsequently provided with the identical SLK base, these gaps consistently narrow. Most notably, the gap between Llama3.1-70B (84.5\%) and GPT-5 (86.6\%) decreases from 12.8 to 2.1 percentage points, corresponding to a net reduction of 10.7 percentage points ($\Delta\downarrow 10.7$, $p < 0.001$). Similar reductions are observed for Llama3.1-8B versus Qwen3-235B ($\Delta\downarrow 3.5$, $p < 0.05$) and DeepSeek-70B versus DeepSeek V3.1 ($\Delta\downarrow 2.2$, $p < 0.05$). These findings suggest that SLK consistently narrows the performance gap between open-weight and SOTA models across different baseline capability levels under identical knowledge conditions.}

\subsection*{Self-learned knowledge has robust predictability and transferability}

\begin{figure}[htbp]
	\begin{center}
		\includegraphics[width=0.9\textwidth]{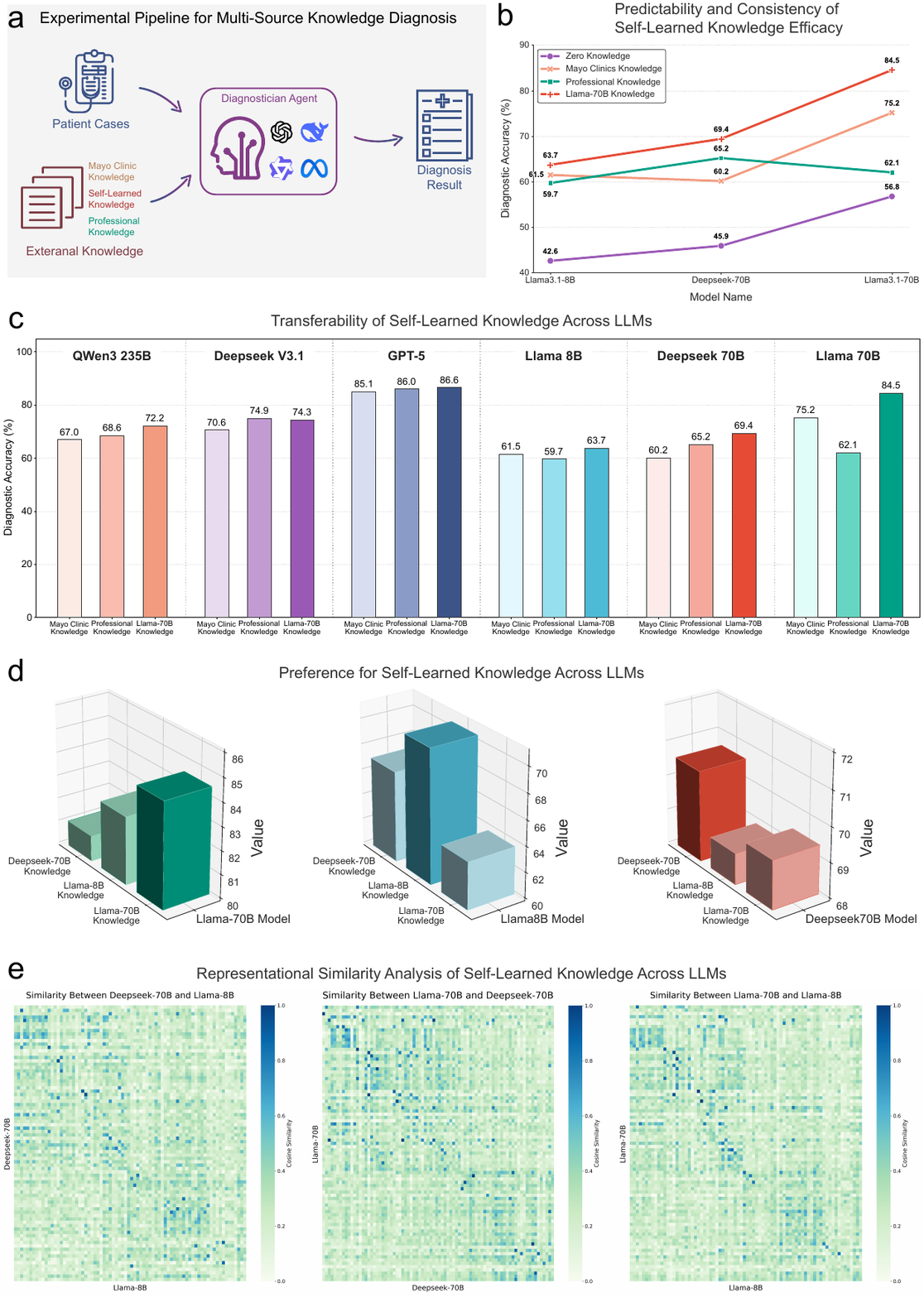}
	\end{center}
	\caption{\textbf{Self-learned knowledge demonstrates a greater stability and cross-model transferability.} (a) Schematic illustration of the experimental pipeline for multi-source knowledge diagnosis. (b) 
    Performance scaling parallels zero-shot baselines, highlighting the predictability of self-learned knowledge efficacy. (c) Cross-model benchmarking demonstrates that self-learned knowledge yields performance gains across diverse LLMs, validating its robust transferability. (d) Analysis reveals a distinct model-specific preference, where models perform optimally using their internally generated knowledge. (e) RSA visualizes distinct model-specific representations while highlighting the structural consistency of diagnostic reasoning for specific diseases.}
	\label{fig:fig3}
\end{figure}

To systematically evaluate the predictability and transferability of Self-Learned Knowledge, we establish a standardized experimental setting in which three distinct knowledge sources, Mayo Clinic Knowledge, Professional Knowledge, and Self-Learned Knowledge, are separately provided to the diagnostician agent for clinical diagnosis (Fig.~\ref{fig:fig3}a). For a controlled comparison, the Self-Learned Knowledge used throughout this analysis is uniformly derived from the highest-performing Llama-3.1-70B agent team.

We first assess the predictability of Self-Learned Knowledge by examining whether diagnostic performance scales consistently with the intrinsic capability of the base model, a desirable property for reliable clinical system deployment.
As shown in Fig.~\ref{fig:fig3}b, the `Zero Knowledge' baseline establishes a clear, linear hierarchy of intrinsic capability, starting with Llama-8B (42.6\%), followed by DeepSeek-70B (45.9\%), and peaking with Llama-70B (56.8\%). However, incorporating authoritative knowledge disrupts this predictability, introducing significant volatility. For instance, the Mayo Clinic Knowledge creates a {V-shaped} performance dip: while it is highly effective for Llama-70B (75.2\%), its efficacy drops sharply for the DeepSeek-70B model (60.2\%), significantly underperforming compared to the smaller Llama-8B (61.5\%). Conversely, the Professional Knowledge exhibits an inverted trend, peaking with DeepSeek-70B (65.2\%) but failing to effectively guide the stronger Llama-70B (62.1\%). In stark contrast, the Self-Learned Knowledge restores clinical predictability. When applying the unified knowledge generated by Llama-70B, the diagnostic accuracy realigns perfectly with the models' intrinsic capability, showing a progressive increase from Llama-8B (63.7\%) to DeepSeek-70B (69.4\%), and culminating at 84.5\% with Llama-70B. This linear progression demonstrates the inherent predictability of Self-Learned Knowledge: as the intrinsic capability of the base model improves, diagnostic accuracy yields a predictable increase, ensuring reliable performance scaling without the risk of unexpected degradation.

Beyond predictability, we assess whether Self-Learned Knowledge can transfer across LLMs with different scales and architectures (Fig.~\ref{fig:fig3}c, Supplementary Table 4). We define transferability as the principle that sufficient knowledge should yield significant performance enhancements across a diverse range of models. Consistent with this definition, the Self-Learned Knowledge distilled by the Llama-70B agent team enhances performance not only for the Llama series but also for entirely different architectures. Notably, the Self-Learned Knowledge universally outperformed the Mayo Clinic Knowledge across all tested models. When compared to the Professional Knowledge, the Self-Learned Knowledge exhibits competitive or superior performance across the board. For the Qwen3-235B model, utilizing Self-Learned Knowledge (72.2\%) yielded a clear advantage over Professional Knowledge (68.6\%). Similarly, for DeepSeek V3.1, Self-Learned Knowledge achieves performance comparable to authoritative knowledge. Even for the SOTA GPT-5, our approach maintains a competitive edge (86.6\%), slightly surpassing Professional Knowledge (86.0\%). This confirms that the knowledge from our framework is not model-specific overfitting but a universal representation of clinical logic, capable of enhancing the diagnostic competence of models ranging from lightweight open-weight to top-tier ones.
To further corroborate the clinical validity of these machine-generated insights, we conduct a human expert evaluation (Table~\ref{tab:pathology_scores_average_plain} and \nameref{Reader-Study} section in \nameref{meth}). Based on the conceptual relevance defined in Supplementary Table 2, experts assess the relevance between the knowledge and specific target diseases. The results indicate that the Self-Learned Knowledge consistently maintains a significant clinical correlation with target diseases, thereby effectively substantiating its medical value and reliability in practical diagnostic scenarios.

\begin{table}[htbp]
\centering
\caption{Evaluation Results (mean ± standard deviation) of the Concept Relevance by Human Experts}
\label{tab:pathology_scores_average_plain}
\begin{tabularx}{\textwidth}{@{} c *{7}{C} @{}}
\toprule
\multirow{2}{*}{Model} & \multicolumn{7}{c}{Pathology} \\
\cmidrule(lr){2-8}
& \makecell{Appendi-\\citis} & \makecell{Cholecy-\\stitis} & \makecell{Diverti-\\culitis} & \makecell{Pancrea-\\titis} & \makecell{Peri-\\carditis} & \makecell{Pneumonia} & \makecell{Pulmonary\\embolism} \\
\midrule

\makecell{Llama\\-8B}  & $3.73\pm0.16$ & $3.26\pm0.51$ & $3.52\pm0.58$ & $3.75\pm0.22$ & $2.92\pm0.66$ & $3.62\pm0.54$ & $3.10\pm0.35$\\
\midrule
\makecell{DeepSeek\\-70B} & $3.95\pm0.16$ & $\textbf{3.89}\pm0.26$ & $3.81\pm0.34$ & $\textbf{4.17}\pm0.09$ & $\textbf{3.74}\pm0.52$ & $3.72\pm0.57$ & $3.62\pm0.51$ \\
\midrule
\makecell{Llama\\-70B} & $\textbf{4.21}\pm0.15$ & $3.77\pm0.21$ & $\textbf{4.03}\pm0.42$ & $\textbf{4.17}\pm0.34$ & $3.28\pm0.33$ & $\textbf{4.00}\pm0.47$ & $\textbf{4.08}\pm0.43$ \\
\bottomrule
\end{tabularx}
\end{table}

\subsection*{LLM prefers knowledge learned by itself}

Our investigation uncovers a distinct ``Self-Preference'' phenomenon: every diagnostician agent achieves peak performance exclusively when utilizing knowledge generated by its native agent team (Fig.~\ref{fig:fig3}d, Supplementary Table 5).
This self-preference is universal and bidirectional. On one hand, replacing a smaller model's knowledge with that from the superior Llama-70B results in performance degradation rather than enhancement; the Llama-8B agent suffers a 7.2 percentage points accuracy drop (from 70.9\% to 63.7\%) when switching to Llama-70B's knowledge. Crucially, this rule applies equally to the most capable model: Llama-70B also fails to maintain its peak performance when utilizing external knowledge. Its accuracy diminishes when guided by insights from DeepSeek-70B or Llama-8B, confirming that diagnostic capability relies not merely on the absolute quality of knowledge, but on the intrinsic compatibility between the knowledge generator and the consumer.
We attribute this to distinct ``cognitive styles'' inherent to each LLM. Representational similarity analysis (heatmap in Fig.~\ref{fig:fig3}e) and linguistic reviews (Supplementary Tables 10, 11, and 12) confirm significant semantic disparities between the knowledge representations of different models. This phenomenon mirrors clinical practice, where physicians develop personalized heuristic understandings of universal medical knowledge; such individualized cognitive schemas are highly effective for the originator but often less transferable to colleagues.

\subsection*{Systematic ablation reveals how context, knowledge, and refinement drive MACD performance}

We conduct systematic ablation studies on four dimensions: context length, knowledge composition, patient information modality, and the Refiner Agent, to identify the key determinants of MACD performance.

Context length shows a non-monotonic effect (Table~\ref{tab:comprehensive_ablation}, Panel A).  The results indicate that performance does not correlate linearly with context length. An 8k token length consistently yields optimal results across most models. For instance, DeepSeek 70B achieves a peak average accuracy of 0.715 at 8k, distinctly outperforming the 0.677 observed at both 4k and 16k lengths. Similarly, Llama 8B reaches its maximum at 0.709 (8k), compared to 0.694 (4k) and 0.694 (16k). Although the context length of 8k (0.845) did not achieve optimal results on Llama-70B, it achieved almost the same results as the lengths of 4k (0.848) and 16k (0.850). These findings confirm that the 8k token length balances sufficient diagnostic detail against the noise introduced by excessive lengths.

Knowledge composition critically influences performance, with general and rare knowledge acting in a complementary manner (Table~\ref{tab:comprehensive_ablation}, Panel B). Using the full self-learned knowledge base yields the highest accuracy across models. Llama 70B reaches 0.845 with full knowledge, exceeding “General” (0.795) and “Rare” (0.752) alone. For DeepSeek 70B, restricting inputs to “Rare” knowledge reduces accuracy to 0.602, while restoring the full set recovers performance to 0.715.

Diagnostic accuracy degrades substantially when clinically important information is removed (Table~\ref{tab:comprehensive_ablation}, Panel C). Models using complete patient information are almost superior to those with partial inputs. Limiting inputs to physical examinations and  { laboratory results} reduces Llama 8B accuracy from 0.709 to 0.598 and DeepSeek 70B from 0.715 to 0.631, underscoring the necessity of integrating heterogeneous clinical evidence.

The Refiner Agent is essential for robust diagnosis by eliminating noise and knowledge redundancy (Panel D). 
While the Summarizer Agent aggregates initial insights, it often retains detrimental noise. The Refiner Agent eliminates this interference and removes knowledge redundancy, significantly enhancing accuracy (e.g., Llama-8B on appendicitis: from 0.595 to 0.895). This demonstrates that refining raw summaries is crucial for ensuring knowledge quality and unlocking effective diagnosis.

\begin{table}[htbp]
  \centering
  \caption{Systematic ablation reveals the contributions of context scale, knowledge integration, patient modality, and refinement to MACD diagnostic performance.}
  \label{tab:comprehensive_ablation}

  \footnotesize
  \setlength{\tabcolsep}{2pt}
  \renewcommand{\arraystretch}{1.1}

  \begin{tabular}{cccccccccc}
    \toprule
    \textbf{Model} & \textbf{Setting} & \shortstack{\textbf{Appendi-}\\\textbf{citis}} & \shortstack{\textbf{Cholecy-}\\\textbf{stitis}} & \shortstack{\textbf{Diverti-}\\\textbf{culitis}} & \shortstack{\textbf{Pancrea-}\\\textbf{titis}} & \shortstack{\textbf{Peri-}\\\textbf{carditis}} & \textbf{Pneumonia} & \shortstack{\textbf{Pulmonary}\\\textbf{embolism}} & \textbf{Average} \\
    
    \midrule
    \multicolumn{10}{l}{\textit{\textbf{Panel A: Impact of Context Length}}} \\
    \midrule
    \multirow{3}{*}{\makecell{Llama\\8B}} 
      & 4k & 0.869 & 0.860 & 0.537 & 0.845 & 0.842 & 0.119 & 0.787 & 0.694 \\
      & 8k & 0.895 & 0.810 & 0.609 & 0.870 & 0.736 & 0.248 & 0.795 & 0.709 \\
      & 16k & 0.876 & 0.857 & 0.553 & 0.845 & 0.825 & 0.119 & 0.784 & 0.694\\
    \cmidrule(l){2-10}
    \multirow{3}{*}{\makecell{DeepSeek\\70B}} 
      & 4k & 0.987 & 0.949 & 0.333 & 0.620 & 0.754 & 0.214 & 0.884 & 0.677 \\
      & 8k & 0.988 & 0.944 & 0.382 & 0.665 & 0.825 & 0.262 & 0.938 & 0.715 \\
      & 16k & 0.988 & 0.951 & 0.333 & 0.622 & 0.754 & 0.210 & 0.884 & 0.677 \\
    \cmidrule(l){2-10}
    \multirow{3}{*}{\makecell{Llama\\70B}} 
      & 4k & 0.988 & 0.960 & 0.768 & 0.726 & 0.895 & 0.677 & 0.923 & 0.848 \\
      & 8k & 0.995 & 0.960 & 0.772 & 0.730 & 0.895 & 0.635 & 0.932 & 0.845 \\
      & 16k & 0.990 & 0.962 & 0.776 & 0.729 & 0.895 & 0.670 & 0.924 & 0.850 \\

    \midrule
    \multicolumn{10}{l}{\textit{\textbf{Panel B: Impact of Self-Learned Knowledge Composition}}} \\
    \midrule
    \multirow{3}{*}{\makecell{Llama\\8B}} 
      & w/ general & 0.830 & 0.785 & 0.618 & 0.833 & 0.702 & 0.146 & 0.788 & 0.672 \\
      & w/ rare & 0.876 & 0.651 & 0.545 & 0.790 & 0.579 & 0.116 & 0.785 & 0.620\\
      & Full & 0.895 & 0.810 & 0.609 & 0.870 & 0.736 & 0.248 & 0.795 & 0.709 \\
    \cmidrule(l){2-10}
    \multirow{3}{*}{\makecell{DeepSeek\\70B}} 
      & w/ general & 0.980 & 0.951 & 0.402 & 0.678 & 0.825 & 0.214 & 0.890 & 0.705 \\
      & w/ rare & 0.971 & 0.881 & 0.268 & 0.587 & 0.386 & 0.185 & 0.933 & 0.602 \\
      & Full & 0.988 & 0.944 & 0.382 & 0.665 & 0.825 & 0.262 & 0.938 & 0.715 \\
    \cmidrule(l){2-10}
    \multirow{3}{*}{\makecell{Llama\\70B}} 
      & w/ general & 0.978 & 0.981 & 0.634 & 0.620 & 0.772 & 0.692 & 0.887 & 0.795 \\
      & w/ rare & 0.994 & 0.943 & 0.756 & 0.784 & 0.474 & 0.365 & 0.952 & 0.752 \\
      & Full & 0.995 & 0.960 & 0.772 & 0.730 & 0.895 & 0.635 & 0.932 & 0.845 \\

    \midrule
    \multicolumn{10}{l}{\textit{\textbf{Panel C: Impact of Patient Case Modality}}} \\
    \midrule
    \multirow{4}{*}{\makecell{Llama\\8B}} 
      & pl & 0.672 & 0.764 & 0.301 & 0.906 & 0.807 & 0.084 & 0.652 & 0.598 \\
      & pi & 0.895 & 0.864 & 0.626 & 0.602 & 0.859 & 0.101 & 0.777 & 0.675\\
      & li & 0.865 & 0.853 & 0.569 & 0.851 & 0.859 & 0.144 & 0.808 & 0.707\\
      & full & 0.895 & 0.810 & 0.609 & 0.870 & 0.736 & 0.248 & 0.795 & 0.709 \\
    \cmidrule(l){2-10}
    \multirow{4}{*}{\makecell{DeepSeek\\70B}} 
      & pl & 0.872 & 0.775 & 0.309 & 0.796 & 0.754 & 0.121 & 0.790 & 0.631 \\
      & pi & 0.986 & 0.951 & 0.479 & 0.422 & 0.737 & 0.239 & 0.897 & 0.673\\
      & li & 0.989 & 0.953 & 0.398 & 0.647 & 0.684 & 0.219 & 0.913 & 0.686\\
      & full & 0.988 & 0.944 & 0.382 & 0.665 & 0.825 & 0.262 & 0.938 & 0.715 \\
    \cmidrule(l){2-10}
    \multirow{4}{*}{\makecell{Llama\\70B}} 
      & pl & 0.875 & 0.806 & 0.496 & 0.839 & 0.877 & 0.569 & 0.836 & 0.757\\
      & pi & 0.990 & 0.974 & 0.821 & 0.477 & 0.789 & 0.643 & 0.918 & 0.801\\
      & li & 0.990 & 0.975 & 0.826 & 0.734 & 0.825 & 0.655 & 0.931 & 0.848\\
      & full & 0.995 & 0.960 & 0.772 & 0.730 & 0.895 & 0.635 & 0.932 & 0.845 \\

    \midrule
    \multicolumn{10}{l}{\textit{\textbf{Panel D: Impact of Agent Components}}} \\
    \midrule
    \multirow{2}{*}{\makecell{Llama\\8B}} 
      & Sum. Only & 0.595 & 0.806 & 0.617 & 0.818 & 0.578 & 0.158 & 0.759 & 0.619 \\
      & Sum. + Ref. & 0.895 & 0.810 & 0.609 & 0.870 & 0.736 & 0.248 & 0.795 & 0.709 \\
    \cmidrule(l){2-10}
    \multirow{2}{*}{\makecell{DeepSeek\\70B}} 
      & Sum. Only & 0.988 & 0.942 & 0.343 & 0.632 & 0.807 & 0.212 & 0.899 & 0.689 \\
      & Sum. + Ref. & 0.988 & 0.944 & 0.382 & 0.665 & 0.825 & 0.262 & 0.938 & 0.715 \\
    \cmidrule(l){2-10}
    \multirow{2}{*}{\makecell{Llama\\70B}} 
      & Sum. Only & 0.995 & 0.963 & 0.737 & 0.710 & 0.922 & 0.610 & 0.907 & 0.835 \\
      & Sum. + Ref. & 0.995 & 0.960 & 0.772 & 0.730 & 0.895 & 0.635 & 0.932 & 0.845 \\
    \bottomrule
  \end{tabular}
  
  \par 
  \smallskip 
  \begin{minipage}{\linewidth}
    \footnotesize
    \textit{Note.} 
    
    \textbf{w/ general}: with general part of self-learned knowledge; 
    \textbf{w/ rare}: with rare part of self-learned knowledge; 
    \textbf{Full}: with complete self-learned knowledge.
    
    \textbf{pl}: with physical examination and laboratory results; 
    \textbf{pi}: with physical examination and image report; 
    \textbf{li}: with laboratory results and image report;
    \textbf{full}: with all available patient case information.
    
    \textbf{Sum. Only}: ablation using only the Summarizer agent (without iterative refinement);
    \textbf{Sum. + Ref.}: using both Summarizer and Refiner agents (complete MACD framework).
  \end{minipage}
\end{table}

\subsection*{Self-learned knowledge outperforms authoritative knowledge in the MACD-human collaboration workflow}

\begin{figure}[htbp]
	\begin{center}
		\includegraphics[width=0.9\textwidth]{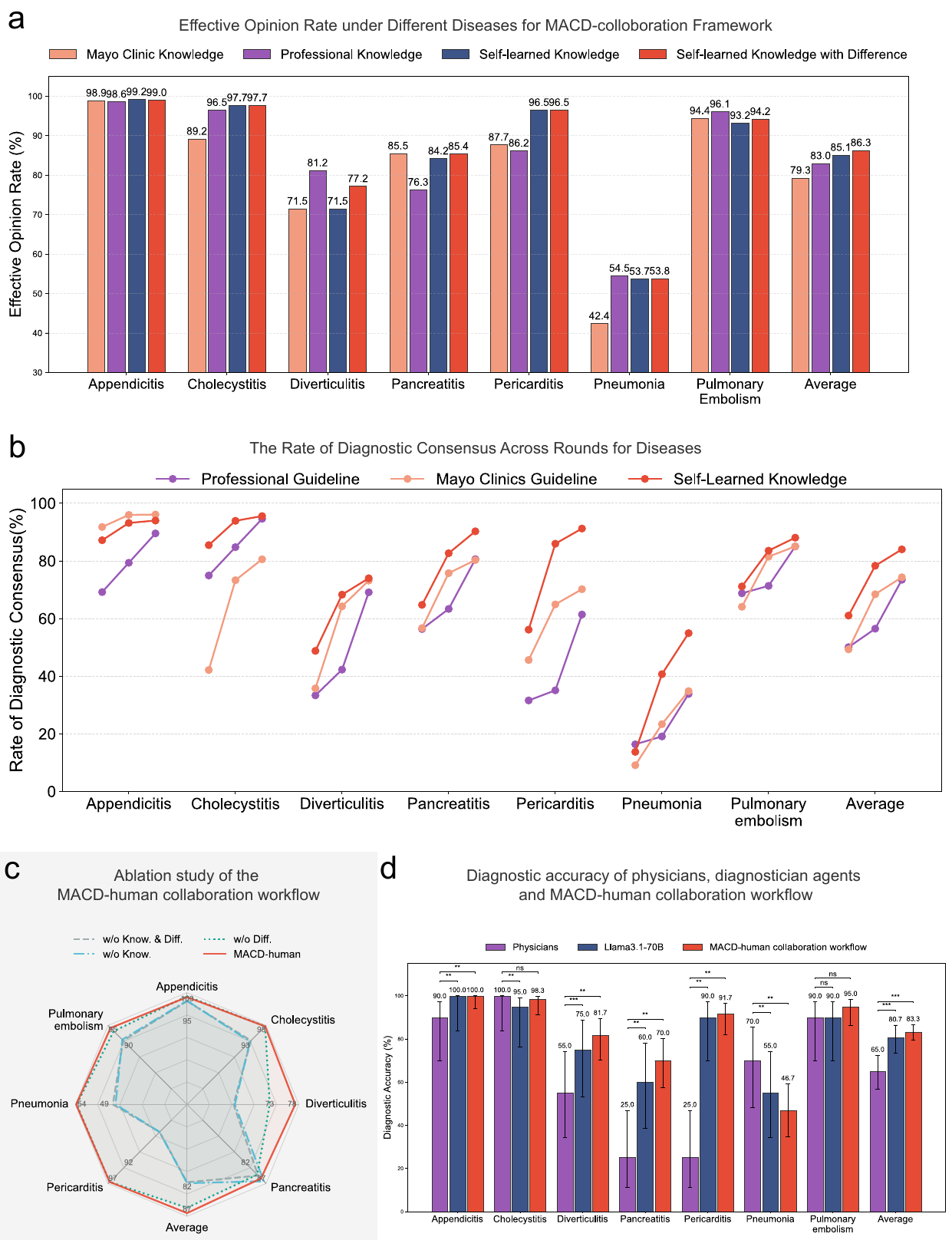}
	\end{center}
	\caption{\textbf{The MACD-human collaboration workflow achieves a higher effective opinion rate with self-learned knowledge.} (a) Performance across knowledge sources. Self-learned Knowledge outperforms external professional knowledge (Mayo Clinic/Professional), and combining it with difference reports yields the optimal effective opinion rate. (b) The proportion of cases with different knowledge types that reach consensus within three rounds of comparison is shown. Self-learned knowledge achieves the highest diagnostic consensus rate. (c) Ablation study of the MACD-human collaboration workflow. Comparisons with variants excluding continuous knowledge or difference reports demonstrate that both components contribute to the overall performance improvement. (d) Comparison of diagnostic accuracy on the MIMIC-MACD-human dataset. The figure illustrates the effectiveness of the workflow in augmenting clinical diagnosis.}
	\label{fig:fig4}
\end{figure}

To further assess the impact of different knowledge sources in a collaborative reasoning context, the MACD-human collaboration workflow is implemented as a medical consultation system to provide effective opinions for human physicians based on the three LLMs above (workflow design detailed in \nameref{MACD-collaboration} of the \nameref{meth} section). We introduce two evaluation metrics: the effective opinion rate and the diagnostic consensus rate; both are defined in \nameref{Evaluation-Metrics}. The framework's performance is evaluated under the guidance of three distinct types of knowledge without a difference report: Self-Learned Knowledge, Mayo Clinic Knowledge, and Professional Knowledge. The results demonstrate that Self-Learned Knowledge holds a significant advantage in the effective opinion rate, providing more valuable reference results for human physicians. The framework utilizing Self-Learned Knowledge with difference reports achieves a superior rate of 86.3\% in Fig.~\ref{fig:fig4}a. This performance is markedly higher than that of the framework guided by Professional Knowledge (83.0\%) and Mayo Clinic Knowledge (79.3\%).

The Self-Learned Knowledge demonstrates the most diagnostic consensus in the workflow. The study analyzes the diagnostic consensus rate to measure how quickly agents converge on a diagnosis. In the initial round of discussion, the framework guided by the Self-Learned Knowledge achieves the highest consensus rate, with a rate of 61.0\%. This surpasses the initial agreements achieved using the Professional Guideline at 50.1\% and the Mayo Clinic Guideline at 49.3\% (Fig.~\ref{fig:fig4}b). This advantage persists through the third round, where the Self-Learned Knowledge reaches 84.0\%, distinctly outperforming the Mayo Clinic Guideline (74.3\%) and the Professional Knowledge (73.5\%).

To evaluate the impact of different modules within the MACD-human collaboration workflow, we conduct an ablation study. The baseline is defined as a setting where self-learned knowledge is provided only during the initial consultation discussion, without the inclusion of difference reports. We compare this baseline against three collaborative settings: providing knowledge in all discussion rounds (w/o Diff.), providing only difference reports (w/o Know.), and providing both knowledge and difference reports (MACD-human).
As illustrated in Fig.~\ref{fig:fig4}c, the results demonstrate that the comprehensive MACD-human workflow achieves optimal performance, attaining an Effective Opinion Rate of 86.3\%. When the difference reports are excluded (w/o Diff.), the rate decreases to 85.1\%, indicating that the difference report contributes to a moderate improvement in consensus quality. In contrast, removing the continuous knowledge provision (w/o Know.) leads to a substantial drop in the effective opinion rate to 79.6\%, a level comparable to the baseline (79.4\%). These findings validate that while difference reports provide detailed clinical analysis to facilitate understanding, the continuous injection of knowledge throughout each discussion round plays a pivotal role in ensuring the effectiveness of medical consultation.

 Furthermore, three human physicians are invited to independently validate the effectiveness of the opinions generated by the workflow on the MIMIC-MACD-human dataset (detailed in \nameref{Reader-Study} in \nameref{meth}). The diagnostic accuracy encompasses both the correct results from cases where the workflow reaches consensus and the final diagnostic outcomes determined by human experts. Fig.~\ref{fig:fig4}d compares the accuracy of the collaboration workflow against the highest-performing single diagnostician agent (Llama3.1-70B) and the human physician cohort  { on text-only vignettes}. The collaboration workflow achieved an accuracy of 83.3\%, compared to 81.0\% for the single Llama3.1-70B agent and 65.0\% for the human physicians. The individual diagnostic accuracy of each physician within the MACD-human workflow is detailed in Supplementary Table 6.

\subsection*{The MACD framework affords explainability in the diagnostic process}

To enhance the transparency and trustworthiness of the diagnostician agents' diagnostic process, the MACD framework incorporates an explainability design that primarily consists of two aspects: causal intervention on Self-Learned Knowledge and the output of the diagnostic rationale.
As part of the knowledge refiner agent's workflow, the study employs a concept-based causal intervention method (\nameref{Multi-Agents} in \nameref{meth}), quantifying the impact of each piece of Self-Learned Knowledge on the final diagnostic accuracy by systematically removing them one by one for a specific disease. Supplementary Tables 7, 8, and 9 present the impact of the knowledge refiner agent based on different models. Notably, the results reveal that removing three concepts with a negative impact on the diagnosis of diverticulitis improves the accuracy of Llama-70B by 3.5 percentage points (detailed in Supplementary Table 9). This method not only facilitates the identification and optimization of the Self-Learned Knowledge content but also provides an explorable basis for understanding the decision-making process of the agents.

Furthermore, the experiment requires the diagnostic agents to explicitly output the diagnostic criteria alongside the final diagnosis. In contrast to `black-box' solutions that only provide a diagnostic result, this textual decision pathway significantly enhances the traceability of the process and the explainability of the final output. Some typical output examples are shown in Supplementary Information S9. For a given case, the MACD framework supports the diagnostic agent with Self-Learned Knowledge while ensuring full transparency by presenting the underlying rationale for each conclusion in the final output. This rationale is closely linked to both the original case content and the provided Self-Learned Knowledge, demonstrating a clear and traceable basis for the diagnostic decision.

\newpage
\section*{Discussions} \label{diss}

The principal finding of this study is that emulating the experience accumulation process of human clinicians offers an effective strategy for narrowing the gap between the broad medical knowledge of LLMs and the nuanced demands of clinical practice. Although prior studies~\cite{hager2024evaluation,chen2025enhancing, bedi2024evaluating, kwon2024large} have consistently shown that LLMs, despite their extensive general medical knowledge, remain limited in diagnostic accuracy and reasoning reliability in complex clinical scenarios, our results demonstrate that MACD can substantially mitigate these limitations through continual accumulation and refinement of case-derived clinical knowledge.

By enabling agents to autonomously construct self-learned knowledge through the cycle of practice, reflection, and knowledge consolidation, the framework achieves an effective translation of general medical knowledge into a self-learned, high-density format. As established by our findings, this mechanism ensures that the resulting knowledge intrinsically matches the model's own reasoning patterns. This cognitive compatibility allows smaller models to utilize knowledge more effectively, enabling them to achieve diagnostic accuracy that rivals or even surpasses much larger systems (e.g., Llama-8B with Self-Learned Knowledge outperforming Qwen-235B with authoritative knowledge). Through this demonstrated ability to substitute parameter scale with knowledge quality, the framework effectively decouples high-performance diagnostic capability from the dependency on massive computational resources. Consequently, MACD offers a strategic, sustainable solution for resource allocation, empowering cost-effective models to deliver SOTA diagnostics in resource-constrained environments.

The superior performance of the MACD framework stems from its ability to simulate the core process of clinical knowledge accumulation, thereby bridging the gap between abstract medical theory and the nuanced reality of clinical practice. First, the framework does not simply force-feed knowledge; instead, it guides agents to autonomously distill and refine concepts from real-world cases, capturing subtleties often omitted from formal guidelines. The Knowledge Summarizer Agent goes beyond extracting high-frequency features found in textbooks, such as \enquote{RLQ pain} for appendicitis and \enquote{Elevated serum lipase level ($>3$ times upper limit)} for pancreatitis. It successfully captures clinically indicative heuristic clues; for instance, the Self-Learned Knowledge for pancreatitis explicitly includes a \enquote{History of recent heavy ethanol consumption} as a potential cause, effectively reinforcing the model’s semantic understanding of key disease features. Following this, the Knowledge Refiner Agent simulates a specialist's reflection. By employing a dual-filter mechanism based on redundancy and importance, it removes misleading entries and anchors the diagnostic agent on key information, creating a highly optimized knowledge repository.
Second, this process reveals a critical insight regarding knowledge format suitability: while human guidelines are authoritative, their structure is often less optimal for LLM processing compared to self-generated content. Although institutional standards serve as the gold standard for clinical theory, their linguistic structure and logical hierarchy are specifically tailored to human cognition and workflows. For example, regarding imaging, guidelines for pancreatitis typically focus on examination modalities and severity grading, while those for diverticulitis concentrate on disease staging based on imaging results. For an LLM-based agent, effectively utilizing such hierarchically complex external information requires extensive additional interpretation, which can introduce comprehension biases. In contrast, the Self-Learned Knowledge incorporates the LLM’s own individualized understanding. It distills these complex concepts into descriptions that are natively compatible with the model’s internal reasoning, such as simplifying detailed imaging criteria into direct diagnostic cues like \enquote{CT scan showing pancreatic inflammation} for pancreatitis. This LLM-centric representation enables the agent to utilize diagnostic insights with seamless application, making the diagnostic process naturally more direct and efficient than navigating the dense structure of human-centric guidelines.

Beyond optimizing performance for the source LLM, a critical finding of this study is that the Self-Learned Knowledge exhibits significant universality and intrinsic authority, effectively establishing a \enquote{LLM-native} knowledge standard. Our investigation into cross-model transferability (Fig.~\ref{fig:fig3}c) reveals that the knowledge distilled by the MACD framework maintains high efficacy even when applied to significantly larger models with distinct architectures, such as Qwen-235B, DeepSeek V3.1, and GPT-5. On these advanced systems, the self-learned knowledge achieves performance parity with, and in some cases surpasses, that obtained using human-authored Professional Guidelines. This robust cross-model applicability confirms that the framework captures fundamental medical truths rather than merely overfitting to a specific model's cognitive bias. It suggests that by autonomously distilling logic from clinical reality, MACD produces a representation of medical knowledge that is universally more accessible and effective for artificial intelligence than traditional human-curated summaries, validating its potential as a reliable, generalized knowledge source for the broader medical AI community.

Compared with existing solutions, such as novel prompt structures or parameter fine-tuning to enhance medical knowledge comprehension~\cite{zhou2024large}, the MACD framework is conceptually distinct and exhibits unique advantages. First, unlike prompting methods that focus on optimizing the reasoning process (e.g., CoT~\cite{wei2022chain}, RoT~\cite{wang2024prompt}, ToT~\cite{yao2023tree}, self-refine~\cite{madaan2023self}), the true value of the framework lies in building a long-term, reusable, and growing accumulation of experience. We do not optimize the model's single inference process; instead, through the accumulation of experience, we continuously narrow the gap between the model's medical knowledge and actual clinical cases, allowing it to learn autonomously in practice in a way that more closely aligns with the growth trajectory of a human expert. Second, compared to solutions that rely on parametric fine-tuning to inject domain knowledge, the framework is more lightweight, flexible, and secure. Fine-tuning methods implicitly and permanently encode knowledge into the model's weights, a process that is costly and difficult to update~\cite{zhou2024large}. In contrast, the framework injects knowledge explicitly and dynamically via prompting. This not only significantly lowers the technical barrier and practical costs but also endows the system with updatability and interpretability, as the accumulated experience can be iterated upon by learning from new, human-readable cases. More importantly, this model allows data processing and updates to be completed locally, providing a feasible solution to the data privacy challenges in medical LLMs.

The technical breakthrough of decoupling diagnostic performance from model scale holds profound implications for optimizing global healthcare resource allocation. By enabling cost-effective, clinically deployable LLMs to achieve expert-level accuracy, the MACD framework offers a viable solution for primary care and grassroots healthcare settings. In resource-scarce regions where specialized medical expertise is limited, such accessible AI agents can function as \enquote{digital specialists} for general practitioners, facilitating early disease identification without the need for immediate referrals to overburdened tertiary centers. Furthermore, in high-demand scenarios like emergency departments, MACD agents can be integrated into intelligent triage workflows. Acting as preliminary diagnostic filters, they can rapidly categorize patient severity and suggest potential diagnoses, allowing human experts to focus their limited attention on the most critical and complex cases. Ultimately, this capability facilitates a more efficient utilization of the existing medical workforce, demonstrating the potential to elevate diagnostic standards in underserved regions and serve as a reliable auxiliary tool for clinical decision making.

Despite the positive results of this study, several limitations remain. First, while the current MACD framework relies on a structured, semi-automated workflow, it proves effective in the diagnostic process. This provides a foundational concept and sets a precedent for the development of more sophisticated, fully automated agent systems. Second, regarding the data, the MIMIC-IV dataset is predominantly text-based. The medical reports within it are written by human physicians, meaning the information is pre-processed and may be subject to human bias. In real-world clinical scenarios, however, physicians supplement their analysis by directly interpreting medical images. Therefore, enabling LLMs to directly process imaging data could potentially enhance their understanding of key patient information. Thirdly, the MIMIC-IV dataset is primarily in English and sourced from the United States. It remains unclear whether the framework can achieve similarly excellent performance on medical data from other countries and regions, and further validation with more diverse clinical data is required.

The method proposed in this study opens up new possibilities for the application of LLMs in clinical diagnosis, and future work can be expanded in the following directions. First, the work evaluates the MACD-human collaboration workflow based on whether at least one agent provides the correct diagnosis, as the results are intended to serve as a reference for human physicians. A deeper investigation is warranted into how the Self-Learned Knowledge can be optimally leveraged within the collaborative consultation setting to maximize the framework's full potential. Additionally, the current MACD framework provides a viable approach for accumulating diagnostic experience. This work demonstrates its feasibility across multiple diseases, showing that knowledge accumulation enhances the model’s diagnostic performance across diverse conditions. The framework holds promise for advancing LLMs’ diagnostic capabilities in specialized disease areas, potentially contributing to more accurate and focused specialty-specific diagnosis in the future. Finally, the medical diagnostic process involves complex ethical and safety issues. The interpretability of current LLMs within this process requires further advancement to achieve a level of trustworthiness acceptable to both clinicians and patients. Also, it is important to note that even though the MACD framework achieves results surpassing human experts in the data, this does not imply its immediate readiness for direct clinical application.

This study successfully constructs and validates the MACD framework. By computationally simulating the experience accumulation process of human physicians, this framework effectively decouples expert-level diagnostic capability from the dependency on massive computational resources. We have demonstrated that identifying a cost-effective pathway to bridge the gap between general medical knowledge and clinical practice is more impactful than merely relying on larger parameter LLMs. More importantly, this work represents a paradigm shift toward sustainable and equitable medical AI. By transforming \enquote{black-box} LLMs into transparent, experience-driven agents that can run on accessible hardware, we provide a promising blueprint for redistributing high-quality medical resources. We believe that this approach will broaden the accessibility of expert-level diagnostics, empowering grassroots healthcare systems and ultimately fostering a more efficient global healthcare ecosystem where quality care is accessible to all, regardless of geographic or economic constraints.

\newpage
\section*{Methods} \label{meth}
\subsection*{Dataset construction} \label{Dataset-Construction}

The study utilizes a composite dataset named MIMIC-MACD, constructed from the MIMIC-IV v2.2 database, comprising a total of 4,390 patient cases across seven distinct abdominal and chest diseases. The dataset integrates two subsets: Abdominal Subset (MIMIC-CDM): Derived from the work of Hager et al.~\cite{hager2024evaluation}, this subset contains 2,400 real-world emergency cases covering four acute abdominal diseases: Appendicitis ($n=957$), Cholecystitis ($n=648$), Diverticulitis ($n=257$), and Pancreatitis ($n=538$). Chest Set: We also construct a new chest disease dataset following identical inclusion principles.  { We filter patients for our targets using the diagnosis table, which contains all recorded diagnostic International Statistical Classification of Diseases and Related Health Problems (ICD) codes~\cite{hager2024evaluation}}. This subset comprises 1,990 cases covering three key chest diseases: Pneumonia ($n=1024$), Pulmonary Embolism ($n=852$), and Pericarditis ($n=114$). All cases follow a standardized structure, retaining de-identified textual information from four key sections: History of Present Illness, Physical Examination, Laboratory Results, and Radiology Reports. Unlike traditional supervised learning, which relies on random data splitting, the MACD framework is designed to simulate a human-like reflection on successful practice. To ensure that the Self-Learned Knowledge is derived from valid clinical reasoning rather than hallucinations or erroneous logic, we implement a success-driven sampling strategy~\cite{zelikman2024star}. This process partitions the dataset into a Learning Set and a Test Set through the following steps: First, the three base diagnostician agents (Llama-3.1-8B, Llama-3.1-70B, and DeepSeek-R1-Distill-Llama3.3-70B) perform  { an initial round of diagnosis} on the entire dataset. For each disease, we identify the subset of cases correctly diagnosed by at least one agent. This filtering ensures that the source material for the knowledge summarization agent contains clear, recognizable diagnostic patterns. From this correct response pool, we randomly sample 90 cases per disease for each agent to serve as its exclusive learning cases. For diseases with high diagnostic difficulty and fewer initial correct diagnoses (specifically Pericarditis), all available correct cases are selected (n=23) to maximize the learning signal. 

Learning Set: The union of all Learning Cases selected by any agent team constitutes the Learning Set. These cases are used solely for the Knowledge Summarizer Agent to generate Self-Learned Knowledge and are strictly excluded from the final evaluation. Test Set: All remaining cases that are not selected for learning constitute the Test Set. Rationale for this partitioning: It is crucial to note that this partitioning strategy inadvertently creates a more challenging evaluation benchmark. Since the Learning Set consists primarily of typical cases that are easily diagnosed in the initial round, the Test Set naturally retains a higher proportion of atypical or complex cases (including those initially misdiagnosed). Evaluating the MACD framework on this residual dataset allows us to rigorously assess its ability to generalize learned patterns to more ambiguous clinical scenarios, providing a robust measure of its real-world utility.

\subsection*{Model selection and deployment}  \label{Model-Selection}

This work selects three general-purpose LLMs as the base models of the MACD framework: Llama3.1-8B, Llama3.1-70B, and DeepSeek-R1-Distill-Llama-3.3-70B. The rationale for their selection is threefold: (1) They all belong to advanced open-source model series and have demonstrated excellent performance on various benchmarks. (2) They differ in parameter scale and training strategies, which provides a basis for evaluating the robustness and effectiveness of the MACD framework on models with varying reasoning capabilities. (3) The models are of a size that is conducive to local deployment.

To ensure the consistency of the experimental results, specific deployment parameters are configured for the models. The context window is limited to 16384 tokens. Additionally, the temperature is set to 0.01, top-k to 1, and top-p to 0.05. All experiments are conducted on NVIDIA A100-80G GPUs.
 { Furthermore, API-based models are integrated for evaluation, including DeepSeek-V3.1 (671B, Deepseek-v3-1-250821), Qwen3 (235B, qwen3-235b-a22b), and GPT-5 (gpt-5-2025-08-07).}

\subsection*{Construction of authoritative knowledge}  \label{Construction-Clinical-Guideline}

To compare the Self-Learned Knowledge autonomously generated by the models with authoritative knowledge, we construct two types of baselines based on clinical guidelines.
We construct the Mayo Clinic Baseline Knowledge by manually extracting relevant diagnostic information for each target disease from the Diseases \& Conditions directory of the official Mayo Clinic Guideline website. For each disease, the complete information from the overview, symptoms, and diagnosis sections is compiled. This extracted content is then directly integrated into the prompt and provided to the LLM as one of the authoritative knowledge sources, as shown in Supplementary Table 13.

We collect clinical practice guidelines published by several authoritative international medical organizations, including the World Society of Emergency Surgery (WSES), the European Society of Cardiology (ESC), and the European Respiratory Society (ERS). Given that the original guideline documents are lengthy and information-dense, we employ the Google Gemini 2.5 Pro (preview) model to process them. Through specific prompt instructions (detailed in Supplementary Information S12), we direct the model to precisely summarize only the diagnostic information relevant to the target diseases, strictly constraining it from adding any of its own knowledge. This process generates highly focused Professional knowledge that remains faithful to the original content, as shown in Supplementary Table 14.

\subsection*{Multi-agent clinical diagnosis framework} \label{Multi-Agents}

\begin{figure}[htbp]
	\begin{center}
		\includegraphics[width=0.9\textwidth]{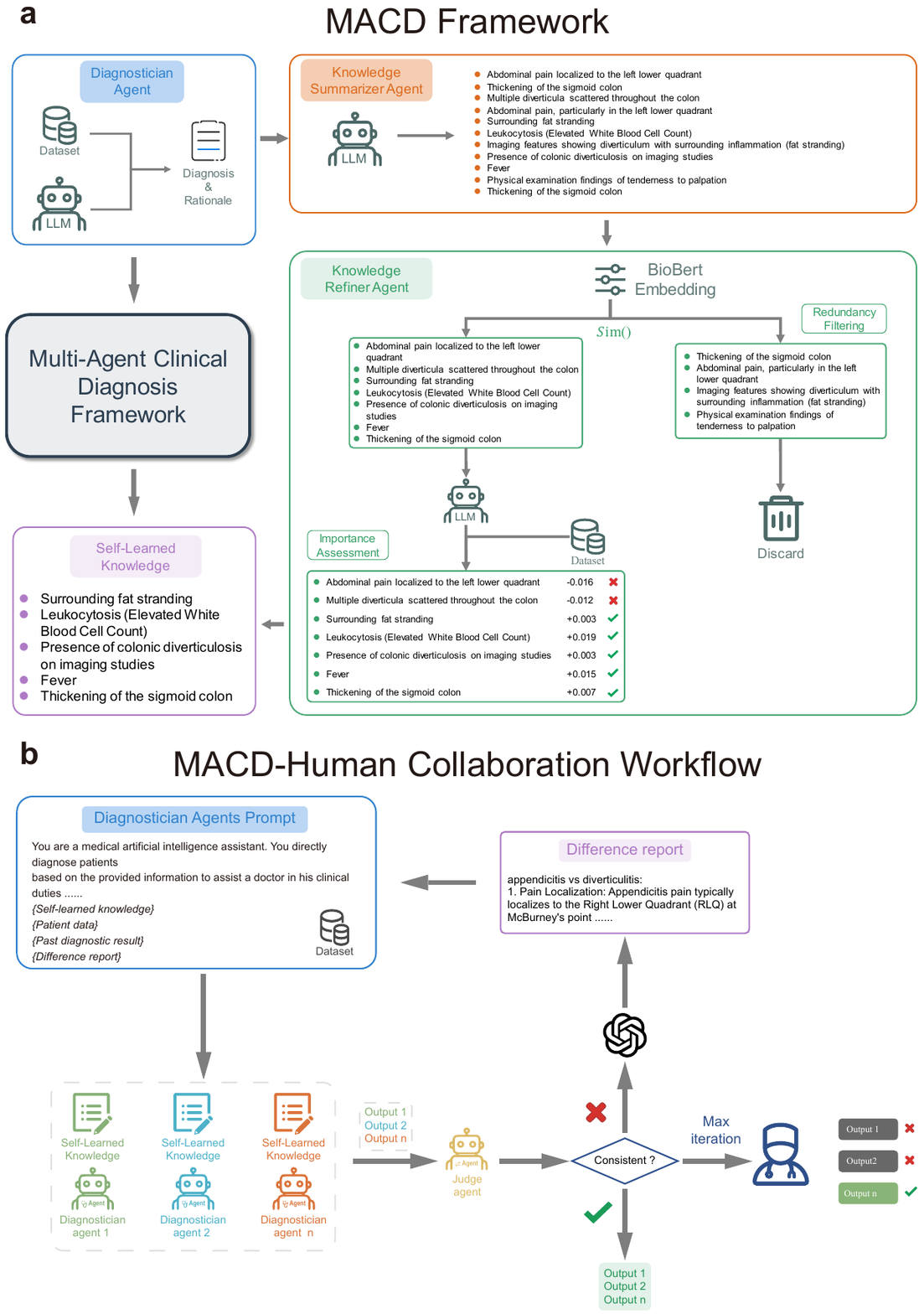}
	\end{center}
	\caption{\textbf{Architectures of the MACD framework and MACD-human collaboration workflow.} (a) The multi-agent clinical diagnosis framework that is composed of diagnostician agents, a knowledge summarizer agent, and a knowledge refiner agent. (b) The MACD-human collaboration workflow is composed of three diagnostician agents, one judge agent, and human physicians. The figure is created by the authors using icons from the built-in Microsoft PowerPoint icon library.}
	\label{fig:fig5}
\end{figure}

The multi-agent clinical diagnosis (MACD) framework comprises two knowledge agents, the Knowledge Summarizer Agent and Knowledge Refiner Agent, and a Diagnostician Agent (Fig.~\ref{fig:fig5}a). The knowledge summarizer agent autonomously reviews learning cases to generate a broad pool of diagnostic concepts for a specific disease.
The knowledge refiner agent then processes this pool, filters it for redundancy and importance, and optimizes Self-Learned Knowledge. For the final diagnosis, this knowledge is provided to either a single diagnostician agent or, for more complex reasoning, to the MACD-human collaboration workflow as a form of internal knowledge when diagnosing new cases. This provides clear, experience-based guidance for the agent’s reasoning process, directing it to focus more precisely on the key features of the disease, thereby improving the accuracy of the final diagnosis.

To formally characterize this process, we formulate MACD as a disease-specific iterative knowledge accumulation and diagnostic procedure. Let $n \in [1, N]$ represent the $n$-th pathology (e.g., appendicitis) and $t \in [1, T]$ denote the $t$-th step of the accumulated knowledge state. In the knowledge accumulation phase, the system evolves through correct diagnostic cases. Initially, the Diagnostician Agent ($\mathcal{M}_{\text{diag}}$) performs diagnoses on a learning dataset $\mathcal{D}_{\text{learning}}^N$ conditioned on the current knowledge base $\mathcal{K}_t^N$:
\begin{equation}\hat{y}_i^N = \mathcal{M}_{\text{diag}}(x_i, \mathcal{K}_t^N), \quad x_i  \in \mathcal{D}_{\text{learning}}^N. \end{equation} 

We filter the generated results to construct a collection of $\hat{y}_i^n$ verified correct diagnoses, denoted as $\mathcal{Y}_{\text{correct}}^N$. Crucially, the Summarizer Agent $\mathcal{M}_{\text{sum}}$ identifies and integrates salient diagnostic insights from this set of diagnoses. This design forces the agent to abstract clinical knowledge from the valid reasoning traces:
\begin{equation}e^N = \mathcal{M}_{\text{sum}}(\mathcal{Y}_{\text{correct}}^N).\end{equation}

The Refiner Agent $\mathcal{M}_{\text{refine}}$ then consolidates and integrates this distilled experience $e^N$ into the knowledge base, updating it $\mathcal{K}_{t+1}^N$ for the next iteration:
\begin{equation}\mathcal{K}_{t+1}^N = \mathcal{M}_{\text{refine}}(\mathcal{K}_t^N, e^N).\end{equation}

In the clinical decision-making phase, the Diagnostician Agent applies the final accumulated knowledge $\mathcal{K}_T = \{ \mathcal{K}_T^1, \mathcal{K}_T^2, \dots, \mathcal{K}_T^N \}$ to unseen cases in the test set $\mathcal{D}_{\text{test}}^N$, ensuring that the clinical deployment benefits from the verified historical experience. Notably, the Diagnostician Agent utilizing this knowledge continues to  { perform diagnosis}. An example of a case study can be found in Supplementary Information S9.
\begin{equation}\hat{y}_j^N = \mathcal{M}_{\text{diag}}(x_j, \mathcal{K}_T), \quad x_j \in \mathcal{D}_{\text{test}}^n, \mathcal{K}_T = \{ \mathcal{K}_T^1, \mathcal{K}_T^2, \dots, \mathcal{K}_T^N \}. \end{equation}

First, as for the diagnostician agent, it performs  { a clinical decision-making task} within the MACD framework. Unlike discriminative multiple-choice classification, the model is not presented with a candidate list to select from. Instead, it operates on a reference-based reasoning paradigm: specific medical knowledge (e.g., Self-Learned Knowledge or Professional Knowledge) is injected into the context window to serve as a diagnostic reference while requiring the agent to actively generate a free-text diagnosis based on logical evidence mapping. For each diagnostic inference, the input sequence is strictly constructed by concatenating three components: A standardized prompt template (Supplementary Information S14) defining the agent's role and output format. The specific medical knowledge context is injected into the context window to act as the conditional anchor for reasoning. The de-identified patient case is structured into four key sections. To ensure rigorous evaluation, each patient case is processed in a completely isolated session with a reset context window, preventing data leakage or cross-sample pattern inference. To effectively address the issue of context overflow, we implement a summarization mechanism when the input length exceeds the maximum context window. An LLM compresses the preceding content into a concise summary before the remaining context is ingested. Furthermore, the agent is mandated to generate dual outputs: the exact primary diagnosis name and the specific supporting criteria extracted from the case. This requirement for explicit evidentiary support serves to mitigate the probability of stochastic guessing, compelling the LLM to engage in a deductive reasoning process prior to generating the final conclusion.

As for the knowledge summarizer agent, we guide the agent to process the key information within the learning cases through carefully designed prompt instructions. We guide the agent to process the key information within the learning cases. In this process, the agent autonomously learns and summarizes clinical insights to form structured diagnostic knowledge, which is categorized into distinct ``General" and ``Rare" components. This knowledge constitutes the agent’s internal representation of the diagnostic logic for a specific disease, its key clinical features, and their associations. The agent’s behavior during this process is constrained by a prompt template structured as a 5-tuple (detailed in Supplementary Information S15). 

Next, as for the knowledge refiner agent, to improve the quality and efficiency of the generated knowledge, we design a dual-filter knowledge refiner agent based on redundancy and importance, detailed as follows. This method first removes repetitive or highly similar diagnostic concepts to preserve semantic diversity. Then, through an importance assessment, it eliminates concepts that could mislead or negatively impact the diagnosis. This dual-filtering process results in a final set of Self-Learned Knowledge that is efficient, reliable, and tailored to the model itself.

The first stage removes redundant concepts that arise because diagnostic features extracted from different cases of the same disease frequently overlap.
We perform a de-duplication operation via BioBERT's semantic embedding method~\cite{deka2022evidence}. Specifically, by calculating the cosine similarity ($Sim(c,c')$) between concept vectors, we employ a greedy algorithm based on maximal marginal relevance to iteratively select candidate concepts that are most semantically distinct from those already chosen,  { resulting in the set of retained diagnostic concepts (top-7)}. This process is applied separately to both the common and rare concept sets to ensure the comprehensiveness of the diagnostic knowledge.

In the second stage, for the concepts retained after redundancy filtering, we implemented a concept-based causal intervention scheme. The retained set may still contain erroneous or spurious concepts that could negatively affect the diagnostician agent.  { To address this, the knowledge refiner agent sequentially ablates each existing diagnostic concept in all cases of the Learning dataset.} It then measures the change in the diagnostic accuracy of the diagnostician agent for the target disease ($\Delta \mathrm{Acc}$) when the concept is present versus absent, thereby capturing the impact of each concept on the outcome. If removing a concept causes accuracy to drop, it is labeled a positive concept; if accuracy improves or remains the same, it is a negative concept. We remove the negative concepts with the most significant adverse impact. The remaining concepts then form the final Self-Learned Knowledge for each target disease (Self-Learned Knowledge can be found in Supplementary Tables 10, 11, and 12, and the difference in semantic similarity is illustrated in Supplementary Figure 2).

Finally, the filtered Self-Learned Knowledge is injected into the prompt template of the diagnostician agent. The model receives this knowledge as prior information before accessing the patient's case data and then proceeds with the diagnosis, thereby enhancing its diagnostic accuracy.

\subsection*{MACD-human collaboration workflow} \label{MACD-collaboration}
To emulate the collaborative nature of real-world clinical consultation, we develop a MACD-human collaboration workflow (Fig.~\ref{fig:fig5}b). This architecture is built upon three distinct diagnostician agents and one judge agent, simulating the process by which multiple physicians conduct a consultation and reach a consensus diagnosis. The framework is designed to function as a medical assistant, providing effective opinions for a human physician's actual diagnostic process. Through a simulated process of consultation and discussion, the three diagnostician agents analyze a case, collect each other's opinions for further analysis, and iterate until a consensus is reached or a maximum number of discussions is met (the prompt template is provided in Supplementary Information S17). The judge agent assesses whether the three diagnostician agents have reached an agreement after each round of discussion. A foundational principle of this framework is that each agent operates using its own unique, self-generated Self-Learned Knowledge, thereby preserving the model-specific knowledge individuality identified in the experiments.

The diagnostic process is governed by the judge agent using a collaborative and iterative strategy detailed in the `Evaluation Metrics'. In each discussion, the three diagnostician agents first diagnose a case independently, and their results are then assessed by the judge agent. If a consensus is not reached, the judge agent invokes GPT-4.1 to generate a difference report that highlights the objective clinical features among the inconsistent results (the prompt template is provided in Supplementary Information S18). The case, along with self-learned knowledge, all conflicting results, and this newly generated difference report, automatically proceeds to the next diagnostic loop. In each subsequent discussion, all diagnostician agents are provided with an enriched informational context, which includes the anonymized diagnostic outputs from the previous round and the difference report to guide their refined analysis. This loop continues until a consensus is reached or a predefined maximum number of discussions is met. For cases that achieve consensus, the agreed-upon result serves as the final output. If no consensus is reached, the diagnostic results from the final round are provided as the ultimate effective opinions; these diagnostic results are then submitted to human physicians for the final diagnosis.

\subsection*{Evaluation metrics} \label{Evaluation-Metrics}

We adopt a hierarchical evaluation strategy to quantitatively analyze diagnostic accuracy. For single-agent performance, we apply a two-level matching system. At the first level, a diagnosis is considered correct if the model's output contains the core medical term for the target disease (exact term matching), ignoring any modifying adjectives. To account for common terminological variations found in clinical practice, we design a secondary, more tolerant matching rule based on anatomical location. This rule accepts reasonable inferences within the same organ system (e.g., accepting ``pericard effusion" as a tolerantly accurate diagnosis for ``pericarditis") and permits modifying adjectives related to disease severity (Supplementary Information S19). Building on the work of Hager et al.\cite{hager2024evaluation}, we add tolerant matching rules for pericarditis, pneumonia, and pulmonary embolism, all of which are reviewed and approved by human experts. 

To evaluate the performance of the MACD framework across the unevenly distributed dataset, we report accuracy at two distinct levels: First, we compute the accuracy individually for each of the seven disease categories, defined as the proportion of correctly diagnosed cases relative to the total cases within the corresponding category ($N_{correct}/N_{total}$ per disease category). To provide a summary metric for the overall system, the average accuracy is reported, which is calculated as the unweighted mean of the seven disease-specific accuracies. This metric treats each disease category as a distinct unit for calculation, independent of its sample size. Each experiment is repeated independently. The reported disease-specific accuracy represents the mean performance across these runs.

In addition to single-agent accuracy, the study establishes a collaboration evaluation strategy to adjudicate consensus within the MACD collaboration framework. This consensus assessment transcends simple string matching to achieve a more clinically meaningful measure of agreement and comprises a two-stage process. First, all raw diagnostic results from the agents are processed using the tolerant name-matching rules described previously. This step normalizes clinically equivalent terms into a core medical term for the target disease while preserving non-conforming diagnostic terms in their original form. Following this standardization, the diagnoses are subjected to a semantic similarity analysis using the BioBERT model.  { A consensus among all diagnoses is confirmed only if their pairwise cosine similarity score surpasses a predefined threshold of 0.82.} This dual methodology ensures that only diagnoses that are truly semantically aligned are considered concordant, providing a robust and reliable method for adjudicating agreement. Based on this, the Diagnostic Consensus Rate is defined as the proportion of total cases for a specific disease in which all three diagnostic agents achieve a unified diagnosis (following semantic processing) in a given diagnostic round. Finally, to evaluate the clinical utility of the framework's final output, the Effective Opinion Rate is introduced, calculated as the proportion of total cases for which the framework provided an effective opinion. The output for a given case is deemed `effective' only if the correct target disease is mentioned at least once in the diagnostic results provided by the collaborative framework; otherwise, the output is considered `ineffective'.

\subsection*{Reader study} \label{Reader-Study}

To establish a human expert performance baseline, seven Emergency Medicine attending physicians with professional experience ranging from 3 to over 20 years are recruited to participate in the diagnostic study. The physicians are divided into two specialized groups: a group of three physicians is responsible for diagnosing the chest disease cases, and a group of four physicians is responsible for the abdominal disease cases. The cases within each category are distributed equally among the physicians in their respective groups. The evaluation primarily covers the following two aspects.

Human experts are invited to evaluate the medical reliability of the Self-Learned Knowledge. Six of the seven participating physicians independently scored each Self-Learned Knowledge generated by the three different LLMs. The core of the evaluation is to assess the strength of association between each concept in the knowledge and its target disease. A 5-point Likert scale is used for evaluation. Finally, the average score from all physicians for all concepts generated by a single model is calculated to serve as the overall relevance score for that model's knowledge.

To establish a reliable reference standard for LLM performance, we evaluate the diagnostic capabilities of human physicians. The physicians are required to independently diagnose a randomly sampled subset of cases, which consists of 80 abdominal cases (20 for each of the four diseases) and 60 thoracic cases (20 for each of the three diseases). To ensure fairness, the case information  { on text-only vignettes} provided to each physician is identical to that given to the LLMs. The case information includes the History of Present Illness, Physical Examination, Laboratory Results, and Radiology Reports. Based on this information, the physicians are asked to provide their most likely primary diagnosis. To prevent physicians from identifying disease patterns, we include an additional 20 abdominal and 15 chest cases of other related diseases as distractor cases. Consequently, each physician in the respective groups evaluates a total of 100 abdominal or 75 thoracic cases. All cases are provided through a website made by ourselves (Supplementary Figures 3, 4, and 5), with the content and presentation format being identical to that given to the LLMs, and the case order is fully randomized. After all physicians complete their diagnoses independently, we calculate the average diagnostic accuracy across the group to serve as the final human expert performance baseline.

\subsection*{Ethics statement} \label{Ethics-statement}
This study used publicly available, de-identified clinical data from the MIMIC-IV v2.2 database and involved no new patient recruitment, clinical intervention, or local collection of patient data. Participating physicians only provided professional evaluations of de-identified cases and model-generated content. The Science and Technology Ethics Committee of the Suzhou Institute for Advanced Research, University of Science and Technology of China determined that this study was exempt from ethics review (Reference No. MCLLSC-2026-001). The requirement for informed consent was waived because the study used publicly available, de-identified data and involved no collection of identifiable patient information. The study was conducted in accordance with the principles of the Declaration of Helsinki.

\newpage
\section*{Code availability}
The source code is publicly available for non-commercial research purposes at \url{https://github.com/qjdzj/MACD}. All LLM prompts are included in the Supplementary Information, with the prompts used.

\section*{Data availability}
The dataset is openly available at the following link for non-commercial purposes: \url{https://github.com/qjdzj/MACD}.

\section*{Acknowledgements}
We would like to express our gratitude to Xin Li for her contributions to the presentation of experimental data, provision of medical references, and proofreading of article content. We thank Chunjiang Wang for his comments on the description of the Supplementary Information section.

This work is supported by the Natural Science Foundation of China under Grants 62502490, 62402473, and 62271465; the Natural Science Foundation of Jiangsu Province under Grant BK20250496; the Sanming Project of Medicine in Shenzhen under Grant SZSM202411033; the National Key R\&D Program of China under Grant 2025YFC3408300;  the Suzhou Basic Research Program under Grant SYG202338; Jiangsu Funding Program for Excellent Postdoctoral Talent, Natural Science Foundation of Jiangsu Province under Grant BK20255001, and the China Postdoctoral Science Foundation under Grant 2024M763178. The funders had no role in the study design, data collection,
data analysis, data interpretation, preparation of the manuscript,
or decision to submit the manuscript for publication.

\section*{Author contributions}
W.L. led the methodological design, experimental validation, and data analysis and contributed to manuscript preparation. R.Y. contributed to the methodological design and manuscript preparation. X.Z. contributed to experimental validation and analysis. K.Z. conceived the main idea, supervised the methodological development, and contributed to manuscript preparation and revision. S.K.Z. provided supervision and revised the manuscript. W.W. coordinated and organized the physician team and participated in the clinical evaluation. L.C., H.Z., J.Z., J.L., M.L., and W.C. provided clinical expertise and participated in the physician evaluation. Z.J. contributed to writing the Introduction and provided suggestions for the revision of the Results. All authors reviewed and revised the manuscript and approved the final version for submission. W.L., R.Y., and X.Z. contributed equally to this work.

\section*{Competing interests}
The authors declare no competing financial or non-financial interests. S. Kevin Zhou is an Associate Editor of npj Digital Medicine. S. Kevin Zhou was not involved in the journal’s review of, or decisions related to, this manuscript. The other authors do not have a competing interest.

\newpage
\bibliography{main}

@article{chang2024survey,
  title={A survey on evaluation of large language models},
  author={Chang, Yupeng and Wang, Xu and Wang, Jindong and Wu, Yuan and Yang, Linyi and Zhu, Kaijie and Chen, Hao and Yi, Xiaoyuan and Wang, Cunxiang and Wang, Yidong and others},
  journal={ACM Transactions on Intelligent Systems and Technology},
  volume={15},
  number={3},
  pages={1--45},
  year={2024},
  publisher={ACM New York, NY}
}

@article{singhal2025toward,
  title={Toward expert-level medical question answering with large language models},
  author={Singhal, Karan and Tu, Tao and Gottweis, Juraj and Sayres, Rory and Wulczyn, Ellery and Amin, Mohamed and Hou, Le and Clark, Kevin and Pfohl, Stephen R and Cole-Lewis, Heather and others},
  journal={Nature Medicine},
  pages={1--8},
  year={2025},
  publisher={Nature Publishing Group US New York}
}

@article{thirunavukarasu2023large,
  title={Large language models in medicine},
  author={Thirunavukarasu, Arun James and Ting, Darren Shu Jeng and Elangovan, Kabilan and Gutierrez, Laura and Tan, Ting Fang and Ting, Daniel Shu Wei},
  journal={Nature Medicine},
  volume={29},
  number={8},
  pages={1930--1940},
  year={2023},
  publisher={Nature Publishing Group US New York}
}

@article{jin2019pubmedqa,
  title={Pubmedqa: A dataset for biomedical research question answering},
  author={Jin, Qiao and Dhingra, Bhuwan and Liu, Zhengping and Cohen, William W and Lu, Xinghua},
  journal={arXiv preprint arXiv:1909.06146},
  year={2019}
}

@article{jin2021disease,
  title={What disease does this patient have? a large-scale open domain question answering dataset from medical exams},
  author={Jin, Di and Pan, Eileen and Oufattole, Nassim and Weng, Wei-Hung and Fang, Hanyi and Szolovits, Peter},
  journal={Applied Sciences},
  volume={11},
  number={14},
  pages={6421},
  year={2021},
  publisher={MDPI}
}

@article{hager2024evaluation,
  title={Evaluation and mitigation of the limitations of large language models in clinical decision-making},
  author={Hager, Paul and Jungmann, Friederike and Holland, Robbie and Bhagat, Kunal and Hubrecht, Inga and Knauer, Manuel and Vielhauer, Jakob and Makowski, Marcus and Braren, Rickmer and Kaissis, Georgios and others},
  journal={Nature Medicine},
  volume={30},
  number={9},
  pages={2613--2622},
  year={2024},
  publisher={Nature Publishing Group US New York}
}

@article{bedi2024systematic,
  title={A systematic review of testing and evaluation of healthcare applications of large language models (LLMs)},
  author={Bedi, Suhana and Liu, Yutong and Orr-Ewing, Lucy and Dash, Dev and Koyejo, Sanmi and Callahan, Alison and Fries, Jason A and Wornow, Michael and Swaminathan, Akshay and Lehmann, Lisa Soleymani and others},
  journal={medRxiv},
  pages={2024--04},
  year={2024},
  publisher={Cold Spring Harbor Laboratory Press}
}

@article{goldberger2000physiobank,
  title={PhysioBank, PhysioToolkit, and PhysioNet: components of a new research resource for complex physiologic signals},
  author={Goldberger, Ary L and Amaral, Luis AN and Glass, Leon and Hausdorff, Jeffrey M and Ivanov, Plamen Ch and Mark, Roger G and Mietus, Joseph E and Moody, George B and Peng, Chung-Kang and Stanley, H Eugene},
  journal={Circulation},
  volume={101},
  number={23},
  pages={e215--e220},
  year={2000},
  publisher={Lippincott Williams \& Wilkins}
}

@inproceedings{zhao2024effective,
  title={Effective In-Context Learning for Named Entity Recognition},
  author={Zhao, Jin and Guo, Qian and Liang, Jiaqing and Li, Zhixu and Xiao, Yanghua},
  booktitle={2024 IEEE International Conference on Bioinformatics and Biomedicine (BIBM)},
  pages={1376--1382},
  year={2024},
  organization={IEEE}
}

@article{sahoo2024systematic,
  title={A systematic survey of prompt engineering in large language models: Techniques and applications},
  author={Sahoo, Pranab and Singh, Ayush Kumar and Saha, Sriparna and Jain, Vinija and Mondal, Samrat and Chadha, Aman},
  journal={arXiv preprint arXiv:2402.07927},
  year={2024}
}

@article{savage2024diagnostic,
  title={Diagnostic reasoning prompts reveal the potential for large language model interpretability in medicine. NPJ Digital Medicine, 7 (1), 20},
  author={Savage, Thomas and Nayak, Ashwin and Gallo, Robert and Rangan, Ekanath and Chen, Jonathan H},
  journal={npj Digital Medicine},
  year={2024}
}

@article{guo2025structured,
  title={Structured Outputs Enable General-Purpose LLMs to be Medical Experts},
  author={Guo, Guangfu and Zhang, Kai and Hoo, Bryan and Cai, Yujun and Lu, Xiaoqian and Peng, Nanyun and Wang, Yiwei},
  journal={arXiv preprint arXiv:2503.03194},
  year={2025}
}

@article{madaan2023self,
  title={Self-refine: Iterative refinement with self-feedback},
  author={Madaan, Aman and Tandon, Niket and Gupta, Prakhar and Hallinan, Skyler and Gao, Luyu and Wiegreffe, Sarah and Alon, Uri and Dziri, Nouha and Prabhumoye, Shrimai and Yang, Yiming and others},
  journal={Advances in Neural Information Processing Systems},
  volume={36},
  pages={46534--46594},
  year={2023}
}

@misc{noauthor_appendicitis_nodate,
	title = {Appendicitis - {Symptoms} and causes},
	url = {https://www.mayoclinic.org/diseases-conditions/appendicitis/symptoms-causes/syc-20369543},
	language = {en},
	urldate = {2025-06-11},
	journal = {Mayo Clinic},
}

@misc{noauthor_cholecystitis_nodate,
	title = {Cholecystitis - {Symptoms} and causes},
	url = {https://www.mayoclinic.org/diseases-conditions/cholecystitis/symptoms-causes/syc-20364867},
	language = {en},
	urldate = {2025-06-11},
	journal = {Mayo Clinic},
}

@misc{noauthor_diverticulitis_nodate,
	title = {Diverticulitis - {Symptoms} and causes},
	url = {https://www.mayoclinic.org/diseases-conditions/diverticulitis/symptoms-causes/syc-20371758},
	language = {en},
	urldate = {2025-06-11},
	journal = {Mayo Clinic},
}

@misc{noauthor_pancreatitis_nodate,
	title = {Pancreatitis - {Symptoms} and causes},
	url = {https://www.mayoclinic.org/diseases-conditions/pancreatitis/symptoms-causes/syc-20360227},
	language = {en},
	urldate = {2025-06-11},
	journal = {Mayo Clinic},
}

@misc{noauthor_pericarditis_nodate,
	title = {Pericarditis - {Symptoms} and causes},
	url = {https://www.mayoclinic.org/diseases-conditions/pericarditis/symptoms-causes/syc-20352510},
	language = {en},
	urldate = {2025-06-11},
	journal = {Mayo Clinic},
}

@misc{noauthor_pneumonia_nodate,
	title = {Pneumonia - {Symptoms} and causes},
	url = {https://www.mayoclinic.org/diseases-conditions/pneumonia/symptoms-causes/syc-20354204},
	language = {en},
	urldate = {2025-06-11},
	journal = {Mayo Clinic},
}

@misc{noauthor_pulmonary_nodate,
	title = {Pulmonary embolism - {Symptoms} and causes},
	url = {https://www.mayoclinic.org/diseases-conditions/pulmonary-embolism/symptoms-causes/syc-20354647},
	language = {en},
	urldate = {2025-06-11},
	journal = {Mayo Clinic},
}

@article{konstantinides20202019,
  title={2019 ESC Guidelines for the diagnosis and management of acute pulmonary embolism developed in collaboration with the European Respiratory Society (ERS) The Task Force for the diagnosis and management of acute pulmonary embolism of the European Society of Cardiology (ESC)},
  author={Konstantinides, Stavros V and Meyer, Guy and Becattini, Cecilia and Bueno, H{\'e}ctor and Geersing, Geert-Jan and Harjola, Veli-Pekka and Huisman, Menno V and Humbert, Marc and Jennings, Catriona Sian and Jim{\'e}nez, David and others},
  journal={European Heart Journal},
  volume={41},
  number={4},
  pages={543--603},
  year={2020},
  publisher={Oxford University Press}
}

@article{metlay2019diagnosis,
  title={Diagnosis and treatment of adults with community-acquired pneumonia. An official clinical practice guideline of the American Thoracic Society and Infectious Diseases Society of America},
  author={Metlay, Joshua P and Waterer, Grant W and Long, Ann C and Anzueto, Antonio and Brozek, Jan and Crothers, Kristina and Cooley, Laura A and Dean, Nathan C and Fine, Michael J and Flanders, Scott A and others},
  journal={American Journal of Respiratory and Critical Care Medicine},
  volume={200},
  number={7},
  pages={e45--e67},
  year={2019},
  publisher={American Thoracic Society}
}

@article{torres2017international,
  title={International ERS/ESICM/ESCMID/ALAT guidelines for the management of hospital-acquired pneumonia and ventilator-associated pneumonia: Guidelines for the management of hospital-acquired pneumonia (HAP)/ventilator-associated pneumonia (VAP) of the European Respiratory Society (ERS), European Society of Intensive Care Medicine (ESICM), European Society of Clinical Microbiology and Infectious Diseases (ESCMID) and Asociaci{\'o}n Latinoamericana del T{\'o}rax (ALAT)},
  author={Torres, Antoni and Niederman, Michael S and Chastre, Jean and Ewig, Santiago and Fernandez-Vandellos, Patricia and Hanberger, Hakan and Kollef, Marin and Bassi, Gianluigi Li and Luna, Carlos M and Martin-Loeches, Ignacio and others},
  journal={European Respiratory Journal},
  volume={50},
  number={3},
  year={2017},
  publisher={European Respiratory Society}
}

@article{asteggiano20152015,
  title={2015 ESC Guidelines for the diagnosis and management of pericardial diseases: The Task Force for the Diagnosis and Management of Pericardial Diseases of the European Society of Cardiology (ESC)},
  author={Adler, Y and Charron, P and Imazio, M and Badano, L and Bar{\'o}n-Esquivias, G and Bogaert, J and Brucato, A and Gueret, P and Klingel, K and Lioni, C and others},
  journal={European Heart Journal},
  volume={36},
  number={42},
  pages={2921–2964},
  year={2015}
}

@article{leppaniemi20192019,
  title={2019 WSES guidelines for the management of severe acute pancreatitis},
  author={Lepp{\"a}niemi, Ari and Tolonen, Matti and Tarasconi, Antonio and Segovia-Lohse, Helmut and Gamberini, Emiliano and Kirkpatrick, Andrew W and Ball, Chad G and Parry, Neil and Sartelli, Massimo and Wolbrink, Daan and others},
  journal={World journal of emergency surgery},
  volume={14},
  pages={1--20},
  year={2019},
  publisher={Springer}
}

@article{sartelli20202020,
  title={2020 update of the WSES guidelines for the management of acute colonic diverticulitis in the emergency setting},
  author={Sartelli, Massimo and Weber, Dieter G and Kluger, Yoram and Ansaloni, Luca and Coccolini, Federico and Abu-Zidan, Fikri and Augustin, Goran and Ben-Ishay, Offir and Biffl, Walter L and Bouliaris, Konstantinos and others},
  journal={World Journal of Emergency Surgery},
  volume={15},
  pages={1--18},
  year={2020},
  publisher={Springer}
}

@article{pisano20202020,
  title={2020 World Society of Emergency Surgery updated guidelines for the diagnosis and treatment of acute calculus cholecystitis},
  author={Pisano, Michele and Allievi, Niccol{\`o} and Gurusamy, Kurinchi and Borzellino, Giuseppe and Cimbanassi, Stefania and Boerna, Djamila and Coccolini, Federico and Tufo, Andrea and Di Martino, Marcello and Leung, Jeffrey and others},
  journal={World journal of emergency surgery},
  volume={15},
  pages={1--26},
  year={2020},
  publisher={Springer}
}

@article{di2016wses,
  title={WSES Jerusalem guidelines for diagnosis and treatment of acute appendicitis},
  author={Di Saverio, Salomone and Birindelli, Arianna and Kelly, Micheal D and Catena, Fausto and Weber, Dieter G and Sartelli, Massimo and Sugrue, Michael and De Moya, Mark and Gomes, Carlos Augusto and Bhangu, Aneel and others},
  journal={World Journal of Emergency Surgery},
  volume={11},
  pages={1--25},
  year={2016},
  publisher={Springer}
}

@article{di2020diagnosis,
  title={Diagnosis and treatment of acute appendicitis: 2020 update of the WSES Jerusalem guidelines},
  author={Di Saverio, Salomone and Podda, Mauro and De Simone, Belinda and Ceresoli, Marco and Augustin, Goran and Gori, Alice and Boermeester, Marja and Sartelli, Massimo and Coccolini, Federico and Tarasconi, Antonio and others},
  journal={World journal of emergency surgery},
  volume={15},
  pages={1--42},
  year={2020},
  publisher={Springer}
}

@article{meta2024introducing,
  title={Introducing llama 3.1: Our most capable models to date, 2024},
  author={Meta, AI},
  journal={URL https://ai. meta. com/blog/meta-llama-3-1/. New models including flagship 405B parameter model, along with upgraded 8B and 70B models featuring 128K context length and multilingual capabilities},
  year={2024}
}

@article{guo2025deepseek,
  title={Deepseek-r1: Incentivizing reasoning capability in llms via reinforcement learning},
  author={Guo, Daya and Yang, Dejian and Zhang, Haowei and Song, Junxiao and Zhang, Ruoyu and Xu, Runxin and Zhu, Qihao and Ma, Shirong and Wang, Peiyi and Bi, Xiao and others},
  journal={arXiv preprint arXiv:2501.12948},
  year={2025}
}

@article{chen2025enhancing,
  title={Enhancing diagnostic capability with multi-agents conversational large language models},
  author={Chen, Xi and Yi, Huahui and You, Mingke and Liu, WeiZhi and Wang, Li and Li, Hairui and Zhang, Xue and Guo, Yingman and Fan, Lei and Chen, Gang and others},
  journal={npj Digital Medicine},
  volume={8},
  number={1},
  pages={159},
  year={2025},
  publisher={Nature Publishing Group UK London}
}

@article{bedi2024evaluating,
  title={Evaluating the clinical benefits of LLMs},
  author={Bedi, Suhana and Jain, Sneha S and Shah, Nigam H},
  journal={Nature Medicine},
  volume={30},
  number={9},
  pages={2409--2410},
  year={2024},
  publisher={Nature Publishing Group US New York}
}

@article{gaber2025evaluating,
  title={Evaluating large language model workflows in clinical decision support for triage and referral and diagnosis},
  author={Gaber, Farieda and Shaik, Maqsood and Allega, Fabio and Bilecz, Agnes Julia and Busch, Felix and Goon, Kelsey and Franke, Vedran and Akalin, Altuna},
  journal={npj Digital Medicine},
  volume={8},
  number={1},
  pages={1--14},
  year={2025},
  publisher={Nature Publishing Group}
}

@article{zhou2024large,
  title={Large language models for disease diagnosis: A scoping review},
  author={Zhou, Shuang and Xu, Zidu and Zhang, Mian and Xu, Chunpu and Guo, Yawen and Zhan, Zaifu and Ding, Sirui and Wang, Jiashuo and Xu, Kaishuai and Fang, Yi and others},
  journal={arXiv preprint arXiv:2409.00097},
  year={2024}
}

@article{wei2022chain,
  title={Chain-of-thought prompting elicits reasoning in large language models},
  author={Wei, Jason and Wang, Xuezhi and Schuurmans, Dale and Bosma, Maarten and Xia, Fei and Chi, Ed and Le, Quoc V and Zhou, Denny and others},
  journal={Advances in Neural Information Processing Systems},
  volume={35},
  pages={24824--24837},
  year={2022}
}

@article{wang2024prompt,
  title={Prompt engineering in consistency and reliability with the evidence-based guideline for LLMs},
  author={Wang, Li and Chen, Xi and Deng, XiangWen and Wen, Hao and You, MingKe and Liu, WeiZhi and Li, Qi and Li, Jian},
  journal={npj Digital Medicine},
  volume={7},
  number={1},
  pages={41},
  year={2024},
  publisher={Nature Publishing Group UK London}
}

@article{yao2023tree,
  title={Tree of thoughts: Deliberate problem solving with large language models},
  author={Yao, Shunyu and Yu, Dian and Zhao, Jeffrey and Shafran, Izhak and Griffiths, Tom and Cao, Yuan and Narasimhan, Karthik},
  journal={Advances in Neural Information Processing systems},
  volume={36},
  pages={11809--11822},
  year={2023}
}

@article{chen2023meditron,
  title={Meditron-70b: Scaling medical pretraining for large language models},
  author={Chen, Zeming and Cano, Alejandro Hern{\'a}ndez and Romanou, Angelika and Bonnet, Antoine and Matoba, Kyle and Salvi, Francesco and Pagliardini, Matteo and Fan, Simin and K{\"o}pf, Andreas and Mohtashami, Amirkeivan and others},
  journal={arXiv preprint arXiv:2311.16079},
  year={2023}
}

@article{chen2024huatuogpt,
  title={Huatuogpt-o1, towards medical complex reasoning with llms},
  author={Chen, Junying and Cai, Zhenyang and Ji, Ke and Wang, Xidong and Liu, Wanlong and Wang, Rongsheng and Hou, Jianye and Wang, Benyou},
  journal={arXiv preprint arXiv:2412.18925},
  year={2024}
}

@inproceedings{kwon2024large,
  title={Large language models are clinical reasoners: Reasoning-aware diagnosis framework with prompt-generated rationales},
  author={Kwon, Taeyoon and Ong, Kai Tzu-iunn and Kang, Dongjin and Moon, Seungjun and Lee, Jeong Ryong and Hwang, Dosik and Sohn, Beomseok and Sim, Yongsik and Lee, Dongha and Yeo, Jinyoung},
  booktitle={Proceedings of the AAAI Conference on Artificial Intelligence},
  volume={38},
  number={16},
  pages={18417--18425},
  year={2024}
}

@inproceedings{deka2022evidence,
  title={Evidence extraction to validate medical claims in fake news detection},
  author={Deka, Pritam and Jurek-Loughrey, Anna and P, Deepak},
  booktitle={International Conference on Health Information Science},
  pages={3--15},
  year={2022},
  organization={Springer}
}

@article{gomez2024artificial,
  title={Artificial-Intelligence-based clinical decision support systems in primary care: A scoping review of current clinical implementations},
  author={Gomez-Cabello, Cesar A and Borna, Sahar and Pressman, Sophia and Haider, Syed Ali and Haider, Clifton R and Forte, Antonio J},
  journal={European Journal of Investigation in Health, Psychology and Education},
  volume={14},
  number={3},
  pages={685--698},
  year={2024},
  publisher={MDPI}
}

@article{delourme2024measured,
  title={Measured performance and healthcare professional perception of large language models used as clinical decision support systems: a scoping review},
  author={Delourme, Sol{\`e}ne and Redjdal, Akram and Bouaud, Jacques and Seroussi, Brigitte},
  journal={Digital Health and Informatics Innovations for Sustainable Health Care Systems},
  pages={841--845},
  year={2024},
  publisher={IOS Press}
}

@article{thomas1996guidelines,
  title={Guidelines in professions allied to medicine},
  author={Thomas, Lois H and Cullum, Nicky A and McColl, Elaine and Rousseau, Nikki and Soutter, Jennifer and Steen, Nick and Cochrane Effective Practice and Organisation of Care Group},
  journal={Cochrane Database of Systematic Reviews},
  volume={2010},
  number={1},
  year={1996},
  publisher={John Wiley \& Sons, Ltd Chichester, UK}
}

@article{custers2015thirty,
  title={Thirty years of illness scripts: theoretical origins and practical applications},
  author={Custers, Eug{\`e}ne JFM},
  journal={Medical Teacher},
  volume={37},
  number={5},
  pages={457--462},
  year={2015},
  publisher={Taylor \& Francis}
}

@inproceedings{zelikman2024star,
  title={Star: Self-taught reasoner bootstrapping reasoning with reasoning},
  author={Zelikman, Eric and Wu, Yuhuai and Mu, Jesse and Goodman, Noah D},
  booktitle={Proc. the 36th International Conference on Neural Information Processing Systems},
  volume={1126},
  year={2024}
}

@article{qiu2025quantifying,
  title={Quantifying the reasoning abilities of LLMs on clinical cases},
  author={Qiu, Pengcheng and Wu, Chaoyi and Liu, Shuyu and Fan, Yanjie and Zhao, Weike and Chen, Zhuoxia and Gu, Hongfei and Peng, Chuanjin and Zhang, Ya and Wang, Yanfeng and others},
  journal={Nature Communications},
  volume={16},
  number={1},
  pages={9799},
  year={2025},
  publisher={Nature Publishing Group UK London}
}

@article{kim2025small,
  title={Small language models learn enhanced reasoning skills from medical textbooks},
  author={Kim, Hyunjae and Hwang, Hyeon and Lee, Jiwoo and Park, Sihyeon and Kim, Dain and Lee, Taewhoo and Yoon, Chanwoong and Sohn, Jiwoong and Park, Jungwoo and Reykhart, Olga and others},
  journal={npj Digital Medicine},
  volume={8},
  number={1},
  pages={240},
  year={2025},
  publisher={Nature Publishing Group UK London}
}
\bibliographystyle{naturemag}

\end{document}